\documentclass[journal=jacsat,manuscript=article]{achemso}

\SectionNumbersOn % So that sections are numbered and we can cross-reference them

\usepackage[utf8]{inputenc} % allow utf-8 input
\usepackage[T1]{fontenc}    % use 8-bit T1 fonts

\usepackage{float}          % float option H
\usepackage{graphicx}       % include graphics
\usepackage[export]{adjustbox}  % align graphics left and right
\usepackage{hyperref}       % hyperlinks
\usepackage{url}            % simple URL typesetting
\usepackage{booktabs}       % professional-quality tables
\usepackage{amsfonts}       % blackboard math symbols
\usepackage{mathtools}      % \coloneqq
\usepackage{xcolor}         % colors
\usepackage{listings}       % code listings
\usepackage{glossaries}
\glsdisablehyper
\usepackage{lscape} 
\usepackage{chngcntr}
\usepackage{verbatim}
\usepackage{tabularx}
\usepackage{array}
\usepackage{multirow}  % \multirow command in tables
\usepackage{makecell}  % \makecell command in tables
\usepackage{mciteplus}
\mciteErrorOnUnknownfalse
\usepackage{array}
\usepackage{ragged2e}
\newcolumntype{P}[1]{>{\RaggedRight\hspace{0pt}}p{#1}}

\newcommand*\dockstring{\textsc{dockstring}}
\newcommand*\compactparagraph[1]{\textbf{#1} \quad}

\newcommand*\figshareurl{\url{https://figshare.com/s/95f2fed733dec170b998}}

\newcommand*\githuburl{\url{https://github.com/dockstring/dockstring}}

\definecolor{codegreen}{rgb}{0,0.6,0}
\definecolor{codegray}{rgb}{0.5,0.5,0.5}
\definecolor{codepurple}{rgb}{0.58,0,0.82}
\definecolor{backcolour}{rgb}{0.95,0.95,0.92}

\lstdefinestyle{mystyle}{
    backgroundcolor=\color{white},   
    commentstyle=\color{codegreen},
    keywordstyle=\color{magenta},
    numberstyle=\tiny\color{codegray},
    stringstyle=\color{codepurple},
    basicstyle=\ttfamily\footnotesize,
    breakatwhitespace=false,         
    breaklines=true,                 
    captionpos=b,                    
    keepspaces=true,                 
    numbers=none,                    
    numbersep=5pt,                  
    showspaces=false,                
    showstringspaces=false,
    showtabs=false,                  
    tabsize=4,
}

\newacronym[firstplural=Gaussian processes (GP)]{gp}{GP}{Gaussian process}
\newacronym{logp}{logP}{log partition coefficient}
\newacronym{qed}{QED}{quantitative estimate of druglikeness}
\newacronym{rl}{RL}{reinforcement learning}
\newacronym{si}{SI}{Supporting Information}

\title{DOCKSTRING: easy molecular docking yields better benchmarks for ligand design}

\author{Miguel García-Ortegón}
\affiliation[University of Cambridge]
{Statistical Laboratory, University of Cambridge, UK}
\email{mg770@cam.ac.uk}
\author{Gregor N.\ C.\ Simm}
\affiliation[University of Cambridge]
{Department of Engineering, University of Cambridge, UK}
\author{Austin J.\ Tripp}
\affiliation[University of Cambridge]
{Department of Engineering, University of Cambridge, UK}
\author{José Miguel Hernández-Lobato}
\affiliation[University of Cambridge]
{Department of Engineering, University of Cambridge, UK}

\author{Andreas Bender}
\affiliation[University of Cambridge]
{Department of Chemistry, University of Cambridge, UK}
\author{Sergio Bacallado}
\affiliation[University of Cambridge]
{Statistical Laboratory, University of Cambridge, UK}
\email{sb2116@cam.ac.uk}

\begin{document}

%%%%%%%%%%%%%%%%%%%%%%%%%%%%%%%%%%%%%%%%%%%%%%%%%%%%%%%%%%%%
% To Do
%%%%%%%%%%%%%%%%%%%%%%%%%%%%%%%%%%%%%%%%%%%%%%%%%%%%%%%%%%%%

% \begin{itemize}
% 	\item Enlarge 2D molecular structures (at least for the molecules that definitely go to the SI) 
% \end{itemize}

%%%%%%%%%%%%%%%%%%%%%%%%%%%%%%%%%%%%%%%%%%%%%%%%%%%%%%%%%%%%
% Abstract
%%%%%%%%%%%%%%%%%%%%%%%%%%%%%%%%%%%%%%%%%%%%%%%%%%%%%%%%%%%%

\begin{abstract}
    % Introduction
    The field of machine learning for drug discovery is witnessing an explosion of novel methods. 
    These methods are often benchmarked on simple physicochemical properties such as solubility or general druglikeness, which can be readily computed. 
    % Problem
    However, these properties are poor representatives of objective functions in drug design, mainly because they do not depend on the candidate's interaction with the target.
    % Possible solution
    By contrast, molecular docking is a widely successful method in drug discovery to estimate binding affinities.
    However, docking simulations require a significant amount of domain knowledge to set up correctly which hampers adoption.
    % DOCKSTRING
    To this end, we present \dockstring{}, a bundle for meaningful and robust comparison of ML models consisting of three components: 
    (1) an open-source Python package for straightforward computation of docking scores; 
    (2) an extensive dataset of docking scores and poses of more than 260K ligands for 58 medically-relevant targets; 
    and (3) a set of pharmaceutically-relevant benchmark tasks including regression, virtual screening, and \textit{de novo} design.
    The Python package implements a robust ligand and target preparation protocol that allows non-experts to obtain meaningful docking scores.
    Our dataset is the first to include docking poses, as well as the first of its size that is a full matrix, thus facilitating experiments in multiobjective optimization and transfer learning.
    %\gncs{The full matrix part needs rewording.}
    % Conclusion
    Overall, our results indicate that docking scores are a more appropriate evaluation objective than simple physicochemical properties, 
    yielding more realistic benchmark tasks and molecular candidates.
\end{abstract}

%%%%%%%%%%%%%%%%%%%%%%%%%%%%%%%%%%%%%%%%%%%%%%%%%%%%%%%%%%%%
\section{Introduction}
\label{sec:introduction}
%%%%%%%%%%%%%%%%%%%%%%%%%%%%%%%%%%%%%%%%%%%%%%%%%%%%%%%%%%%%

% Broad introductory sentence
The field of industrial drug discovery is undergoing a crisis.
Despite significant technological advances, R\&D costs have grown by orders of magnitude while the probability of success of candidate molecules has decreased.
This phenomenon is partly attributed to a lack of sufficiently predictive experimental and computational models \cite{scannell2016when}.
Machine learning (ML) is widely regarded as a promising technology to tackle this issue by providing faster and more accurate models \cite{bender2021artificial}.

% Benchmarks
The rapid development of ML methods for drug discovery\cite{vamathevan2019applications,lavecchia2015machine}
has resulted in a growing need for high-quality benchmarks to allow for these methods to be evaluated and compared.
Ideally, a good benchmark would test a model on accurate experimental data (e.g.\@ experimental bioactivity data)
in a realistic problem setting (e.g.\@ prospective search),
so that strong performance on the benchmark would imply strong performance on real-world tasks.
Unfortunately, the high cost and difficulty of collecting experimental data makes such benchmarks impractical.
Existing benchmarks tend to either 
(1) use a fixed experimental dataset for problem settings like in-distribution regression\cite{wu2018moleculenet}, or
(2) use simple computational properties for problem settings like \textit{de novo} design.
The latter type of benchmark is popular in the ML community
with the tasks of maximizing the \gls{qed}\cite{bickerton2012quantifying}
and penalized \gls{logp}
being highly prevalent.\cite{xu2020reinforced, ahn2020guiding, mollaysa2020goal-directed, samanta2020nevae, maziarka2020mol-CycleGAN, thiede2020curiosity, wu2020bayesian, tripp2020sample}
However, the simplicity of these properties raises doubts about whether performance on such benchmarks is indicative of performance on real drug-design tasks.

% Molecular docking
Previous works have suggested that molecular docking could form the basis for high-quality benchmarks.\cite{coley2020autonomous,Cieplinski2021We,huang2021therapeutics}
% What is it?
Molecular docking is a computational technique that predicts how a small molecule (the \emph{ligand}) binds to a 
protein receptor (the \emph{target}) by simulating the physical interaction between the two.\cite{Varela-Rial2021Structure}
The output of this simulation is a \emph{docking score}, which represents the strength of binding between the molecule and protein, 
and a \emph{docking pose}, the predicted 3D conformation of the ligand in the protein binding pocket.
Below, we summarize some of the benefits of molecular docking over simple physicochemical properties (e.g., logP):
\begin{enumerate}
	\item Interpretability: 
	      docking scores have a clear biological interpretation in terms of binding affinity \cite{kitchen2004docking}, correlating with experimental values in some protein families\cite{su2019comparative}.
	\item Relevance: 
	      docking scores are routinely employed by medicinal chemists in academia and industry to discover hits in virtual screening experiments.
	      Docking poses are also used to identify and exploit important interactions during lead optimization.
	\item Computational cost:
	      docking scores can typically be computed in under a minute,
	      unlike other computational methods like free energy perturbation calculations
	      or density-functional theory.
	\item Challenging benchmark: 
	      the relationship between molecular structure and docking score is complex,
	      as the docking score depends on the 3D structure of the ligand-target complex.
	      Therefore, tasks based on docking require ML models to learn complex 3D features.
\end{enumerate}

% Why do we need a standardized package?
Because of these advantages it is unsurprising that several recent works have applied
ML to tasks based on docking scores.\cite{jeon2020autonomous,graff2021accelerating,Thomas2021Comparison,gentile2020deep,lyu2019ultra}.
Yet, there are still several hurdles which make a docking benchmark difficult to realize.
First, such a benchmark mandates high-quality standardization.
Running a docking simulation involves many intermediate steps, such as target and ligand preparation and the specification of a \textit{search box}.
Each step requires significant domain expertise, and for a benchmark to facilitate a meaningful comparison between algorithms, 
they must be carried out correctly and consistently.
Second, the benchmark needs to be accessible to those without domain knowledge. 
%\gncs{I don't understand the last hurdle. Isn't that more of a requirement?}
Finally, the benchmark needs to contain breadth and diversity of targets.

% Related work
A fully-automated docking software pipeline is a potential way to overcome these hurdles.
Indeed, there are several existing works which try to facilitate the use of molecular docking for ML benchmarks.
However, these works all lack at least one of the aforementioned desiderata.
VirtualFlow \cite{gorgulla2020open} and DockStream \cite{Guo2021DockStream}
(part of the REINVENT ecosystem\cite{Blaschke2020REINVENT})
are general-purpose wrappers for docking programs.
However, they primarily cater docking experts requiring manually-prepared target files and specialized arguments.
The Therapeutics Data Commons (TDC) \cite{huang2021therapeutics} and Cieplinski et al.\cite{Cieplinski2021We}
provide wrappers which offer computation of docking scores from just a SMILES string.
However, both wrappers have shortcomings with respect to standardization.
Neither TDC nor Cieplinski et al. control sources of randomness during the docking procedure
(e.g.,\@ random seeds input into the docking program or the conformer generation routines)
leading to the potential for considerable variance between runs on the same molecule.
Further, at the time of writing, both wrappers have a relatively rudimentary ligand preparation pipeline:
for example, neither of them perform ligand protonation, 
an important part of the preparation process \cite{brink2009influence,bender2021practical}.
Moreover, both of these wrappers provide only a small number of targets:
TDC provides only one target, while Cieplinski et al.\cite{Cieplinski2021We} provide just four.

%\gncs{Are we sure we don't want to mention any dataset?}
% %% Dataset
% Beside wrappers, there also exist several large-scale docking score datasets.
% % DeepDocking: Large dataset and large screening, but not many targets 
% For instance, Gentile et al.\cite{gentile2020deep} presented a novel deep learning platform to predict docking scores of 1.36 billion molecules.
% % Ultra-large library docking for discovering new chemotypes
% Further, Lyu et al.\cite{lyu2019ultra} compute a large dataset of 170 million docking scores against a single target.
% Unfortunately, they do not come with an accessible docking tool to extend the dataset or any novel benchmark tasks.

% Classical benchmarks
In addition to wrappers, several docking benchmarks have been developed.
The Directory of Useful Decoys Enhanced (DUD-E) \cite{mysinger2012directory} is a relatively small ligand set of actives and property-matched decoys for 102 targets.
Originally designed to evaluate docking algorithms, its ligand set has since been widely applied to benchmark ML models for virtual screening \cite{wallach2015atomNet,yan2017proteinLigand,ragoza2017proteinLigand}.
However, it has been argued 
%\gncs{By whom?} 
that simple physicochemical properties (e.g., molecular weight or logP) 
are not sufficient to match actives and decoys against ML algorithms,
%\gncs{I still don't know what matching means here.}
so these can simply memorize the actives and inactives, overfitting and complicating generalization.
Therefore, ML methods employing the ligand set in DUD-E are likely to overestimate virtual screening performance \cite{chen2019hidden,sieg2019in}.
% ML benchmarks
The evaluation framework GuacaMol \cite{brown2019guacamol} provides both a distribution matching and goal-directed benchmark suite,
with the latter containing 20 distinct tasks based on molecular fingerprints,
substructure matching, and physicochemical properties.
Although most of these tasks are challenging, they are largely based on simple physicochemical properties and similarity functions such as the Tanimoto similarity.
As a result, they fail to capture subtleties related to 3D molecular structure or interactions with biomolecules.
%\gncs{I know what you mean but this needs to be worded more precisely.}
The benchmark suite MOSES \cite{polykovskiy2020molecular} provides several molecular generation benchmarks
that focus on generating a diverse set of molecules rather than optimizing for any particular chemical property.

% Overview
In this work, we introduce \dockstring{}, a bundle for standardized and accessible benchmarking of ML models based on molecular docking. 
It consists of three parts: 
a Python package for easy computation of docking scores; 
a large and diverse dataset of docking scores and poses for pre-training; 
and a set of meaningful benchmark tasks on which to evaluate models.
%\gncs{Didn't we say that we want to put the dataset first and then describe the Python package?}
\begin{enumerate}
	\item Python package:
	      a user-friendly Python wrapper of the popular docking package AutoDock Vina \cite{Trott2010AutoDock} (Sections~\ref{sec:wrapper-methods} and \ref{sec:wrapper}).
	      AutoDock Vina was selected due to its high-quality docking poses, reasonable accuracy of predicted binding free energies, 
	      and low computational cost \cite{su2019comparative,Pagadala2017Software}.
	      The emphasis of our package is on simplicity---a docking simulation can be setup in just four lines of code.
	\item Dataset:
	      a dataset of over 260K diverse and druglike molecules docked against a curated list of 58 targets, resulting in more than 15M docking scores and poses (Sections~\ref{sec:dataset-methods} and \ref{sec:dataset}).
	      The high number of activity labels per ligand makes our dataset highly suitable for the design of meaningful benchmark tasks in ML settings such as multi-objective optimization or transfer learning. 
	      Furthermore, targets are selected to represent a number of protein families of high pharmaceutical value, such as kinases or nuclear receptors.
	      Overall, more than 500K CPU hours were invested in the creation of the dataset.
	\item Benchmarks:
	      a set of pharmaceutically-relevant and challenging benchmark tasks covering regression, virtual screening and \emph{de novo} design (Sections~\ref{sec:benchmarks-methods} and \ref{sec:benchmarks}). 
\end{enumerate}

\begin{figure}[ht]
	\centering
	\includegraphics{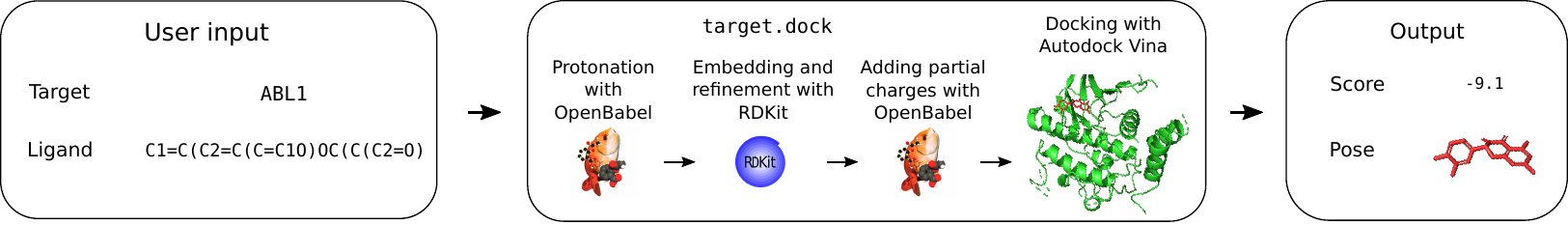}
	\caption{
        Summary of \dockstring{} pipeline from SMILES strings to scores and poses. 
        The method \texttt{target.dock} performs ligand preparation with Open Babel and RDKit, and docking with AutoDock Vina.
	}
	\label{fig:pipeline_summary}
\end{figure}

%%%%%%%%%%%%%%%%%%%%%%%%%%%%%%%%%%%%%%%%%%%%%%%%%%%%%%%%%%%%
\section{Methods}
%%%%%%%%%%%%%%%%%%%%%%%%%%%%%%%%%%%%%%%%%%%%%%%%%%%%%%%%%%%%

%%%%%%%%%%%%%%%%%%%%%%%%%%%%%%%%%%%%%%%%%%%%%%%%%%%%%%%%%%%%
\subsection{Python Package}
\label{sec:wrapper-methods}
%%%%%%%%%%%%%%%%%%%%%%%%%%%%%%%%%%%%%%%%%%%%%%%%%%%%%%%%%%%%

%%%%%%%%%%%%%%%%%%%%%%%%%%%%%%%%%%%%%%%%%%%%%%%%%%%%%%%%%%%%
\subsubsection{Target Preparation}
\label{ssec:target_prep}
%%%%%%%%%%%%%%%%%%%%%%%%%%%%%%%%%%%%%%%%%%%%%%%%%%%%%%%%%%%%

There are 58 prepared targets available in \dockstring{}.
PDB files of 57 protein targets were downloaded from the Directory of Useful Decoys Extended (DUD-E), 
a database of proteins and ligands for benchmarking docking algorithms \cite{mysinger2012directory}.
Structures in DUD-E were determined experimentally to high precision, 
with the large majority of resolutions being less than 2.5~\AA. %and the poorest being 3.3 {\AA}.
Furthermore, DUD-E targets were prepared to improve correlation between theoretical and experimental binding affinities.
For instance, in a few cases, the authors of DUD-E manually added cofactors or crystallographic waters, 
or changed the protonation states of side residues \cite{mysinger2012directory}.
For \dockstring{}, the PDB files were standardized with Open Babel \cite{OBoyle2011Open}
(e.g., the symbols of some metal atoms were not recognized by AutoDock Tools \cite{morris2009autoDock4}),
polar hydrogen atoms were added, and conversion to the PDBQT file format was carried out with AutoDock Tools.

The only target that does not originate from DUD-E is DRD2, the dopamine receptor D2.
It was included in \dockstring{} due to its popularity in molecular regression and optimization \cite{olivecrona2017molecular,arus-Pous2020sMILES-based,maragakis2020a,jin2018learning,jin2019hierarchical}.
To ensure consistency, the preparation of DRD2 was analogous to that of its homolog DRD3 in DUD-E.
Starting from a crystal structure of DRD2 (PDB entry 6CM4) \cite{wang2018structure},
the bound inhibitor (risperidone) as well as residual water and solute molecules were manually removed with PyMOL \cite{PyMOL}, since DRD3 in DUD-E did not include any waters or ions.
Subsequently, the structure was optimized with the program \verb|obminimize| from Open Babel using the general Amber force field (GAFF) \cite{wang2004development}.
Protonation was carried out at pH~7.4 with PROPKA \cite{propka}.
Finally, addition of polar hydrogen atoms and conversion to PDBQT was performed with AutoDock Tools.
%\gncs{What kind of partial charges did you compute?}
%\gncs{I assume these partial charges were also calculated for the 57 other targets?}

The search box of each target in \dockstring{} was also determined.
Every DUD-E structure has a corresponding ligand file from which the box position and size were derived.
We computed the maximum and minimum coordinates of each ligand across each dimension and padded with 12.5~{\AA} on all sides.
Finally, if a box length did not reach 30~{\AA} after padding, we set it to this amount.
The padding length and the minimum box length were tuned manually to minimize the number of positive scores, which indicate highly constrained poses.
The search box of DRD2 was set manually upon visual examination of the binding pocket in the reference structure bound to risperidone \cite{wang2018structure}.

%%%%%%%%%%%%%%%%%%%%%%%%%%%%%%%%%%%%%%%%%%%%%%%%%%%%%%%%%%%%
\subsubsection{Ligand Preparation}
\label{ssec:ligand_prep}
%%%%%%%%%%%%%%%%%%%%%%%%%%%%%%%%%%%%%%%%%%%%%%%%%%%%%%%%%%%%

Ligands are provided to the \dockstring{} package as SMILES strings.
First, \dockstring{} performs a sanity check on the ligand.
Ligands with radicals or ligands consisting of more than one molecular fragment are rejected.
Next, the ligand is (de-)protonated at pH 7.4 with Open Babel \cite{OBoyle2011Open}.
While automated protonation protocols are not perfect \cite{brink2009influence}, their application is widely regarded as good practice \cite{bender2021practical}.
Then, a single three-dimensional (3D) conformation is generated with the Euclidean distance geometry algorithm ETKG \cite{Riniker2015Better}
as implemented in RDKit \cite{RDKit2021033}. This conformation is subsequently refined with the classical force field MMFF94 \cite{Halgren1996Merck}.
During the embedding of the graph representation into a 3D structure, the stereochemistry of determined stereocenters is maintained, whereas any undetermined stereocenters are assigned randomly (but consistently across different runs to ensure the reproducibility of docking scores).
Finally, \dockstring{} computes the Gasteiger charges \cite{Gasteiger1980Iterative} for all atoms and creates a ligand PDBQT file with Open Babel.

%%%%%%%%%%%%%%%%%%%%%%%%%%%%%%%%%%%%%%%%%%%%%%%%%%%%%%%%%%%%
\subsubsection{Molecular Docking with AutoDock Vina}
\label{ssec:docking}
%%%%%%%%%%%%%%%%%%%%%%%%%%%%%%%%%%%%%%%%%%%%%%%%%%%%%%%%%%%%

\dockstring{} docks a ligand against a target using AutoDock Vina \cite{Trott2010AutoDock}. 
The ligand PDBQT input file is obtained automatically as explained in Section~\ref{ssec:ligand_prep}, 
while the target PDBQT and search boxes are taken from the list of prepared input files as explained in Section~\ref{ssec:target_prep}. 
Docking is performed with the default values of exhaustiveness (8), maximum number of binding modes (9) and energy range (3). 
After docking is complete, we obtain up to nine poses, together with their binding free energies.
Note that in subsequent analyses, we only use the lowest docking score (i.e. the best one).

Ligand preparation and molecular docking, and thus the docking score, depend on a random seed. 
We investigated this dependence and found no target-ligand combination for which the docking scores deviated by more than $0.1$~kcal/mol.
Subsequently, we fixed the random seed to obtain a fully deterministic pipeline.

%%%%%%%%%%%%%%%%%%%%%%%%%%%%%%%%%%%%%%%%%%%%%%%%%%%%%%%%%%%%
\subsection{Dataset}
\label{sec:dataset-methods}
%%%%%%%%%%%%%%%%%%%%%%%%%%%%%%%%%%%%%%%%%%%%%%%%%%%%%%%%%%%%

%\gncs{Shouldn't this section go before the Python package?}

%%%%%%%%%%%%%%%%%%%%%%%%%%%%%%%%%%%%%%%%%%%%%%%%%%%%%%%%%%%%
\subsubsection{Target and ligand selection}
%%%%%%%%%%%%%%%%%%%%%%%%%%%%%%%%%%%%%%%%%%%%%%%%%%%%%%%%%%%%

As explained in Section~\ref{ssec:target_prep}, most targets originate from from DUD-E \cite{mysinger2012directory},
a database of proteins and ligands for comparison and development of docking algorithms.
These targets are medically relevant and cover a large variety of protein families, functions, and structures.
We only selected targets with more than 1000 experimental actives in ExCAPE \cite{sun2017exCAPEDB},
a database that curates bioactivity assays from PubChem \cite{kim2021pubChem} and ChEMBL \cite{mendez2019chEMBL},
to ensure a high number of positive examples.
In addition to the targets from DUD-E,
we also included the target DRD2, a popular benchmark in ML \cite{olivecrona2017molecular,arus-Pous2020sMILES-based,maragakis2020a,jin2018learning,jin2019hierarchical}.

% ExCAPE
The ligands and their experimental activity labels originate from ExCAPE \cite{sun2017exCAPEDB}.
We selected all ligands with active labels against the proteins in our target set (at least 1000 actives for each target).
We believe that the large number of positive examples will create a strong signal that will facilitate learning in tasks that require experimental labels.
We also added 150K ligands which only had inactive labels against our targets. 
After discarding 1.8\% of molecules due to failures in the ligand preparation process, 
the final dataset consisted of $260,155$ compounds.

%%%%%%%%%%%%%%%%%%%%%%%%%%%%%%%%%%%%%%%%%%%%%%%%%%%%%%%%%%%%
\subsubsection{Clustering and scaffold analysis}
%%%%%%%%%%%%%%%%%%%%%%%%%%%%%%%%%%%%%%%%%%%%%%%%%%%%%%%%%%%%

Density-Based Spatial Clustering of Applications with Noise (DBSCAN) was implemented with \texttt{scikit-learn} \cite{scikit-learn}.
Prior to executing DBSCAN, molecules were embedded in a space of RDKit fingerprints of path length six and set the neighborhood cutoff $\varepsilon$ to a Jaccard distance of 0.25 \cite{Bender2004Molecular}.
The choice of the fingerprints was motivated by previous analysis by Landrum \cite{landrum_fp_analysis} suggesting that this type of fingerprint was the most appropriate for similarity search.
Bemis-Murcko scaffold decomposition was implemented with RDKit.

%%%%%%%%%%%%%%%%%%%%%%%%%%%%%%%%%%%%%%%%%%%%%%%%%%%%%%%%%%%%
\subsection{Benchmarks}
\label{sec:benchmarks-methods}
%%%%%%%%%%%%%%%%%%%%%%%%%%%%%%%%%%%%%%%%%%%%%%%%%%%%%%%%%%%%

\dockstring{}'s combination of a docking package and large dataset
allows it to underpin a wide variety of benchmark tasks for
supervised learning, active learning, transfer learning, meta-learning, molecule optimization and more.
We formulate benchmark tasks for three problem settings:
regression, virtual screening, and \textit{de novo} design (Table~\ref{tab:benchmarks-summary-table}).
The regression benchmark (Sections~\ref{ssec:regression-benchmark-methods} and \ref{ssec:regression-benchmark}) is relatively standard and widely applicable;
it primarily illustrates the difficulty of predicting docking scores.
Virtual screening (Sections~\ref{ssec:virtual_screening-methods} and \ref{ssec:virtual_screening}) evaluates a model's ability to select active molecules from a large predefined library.
This is a common use case for predictive models in the pharmaceutical industry
and requires strong out-of-distribution performance to be successful.
It is applicable to any method that can rank a list of molecules, either by regression or by other means.
\textit{De novo} design (Sections~\ref{ssec:molopt-methods} and \ref{ssec:molopt}) evaluates the ability to generate novel molecules that optimize an objective function.
It is an active area of research because chemical space is vast (more than $10^{60}$ by some estimates \cite{bohacek1996the}),
so even the largest libraries cover just a tiny fraction of it.
%\gncs{Weird sentence.}
Models for \textit{de novo} design include optimization algorithms, reinforcement learning agents, or generative models.
The objective functions presented here are all based on docking scores but vary in difficulty.

\begin{table}[ht]
	\centering
	\caption{
		Overview of benchmarks tasks in the \dockstring{} bundle.
	}
	\label{tab:benchmarks-summary-table}
	\centering
		\resizebox{\columnwidth}{!}{%
		\begin{tabular}{@{}m{2cm}m{5cm}m{4cm}m{3.5cm}m{11.8cm}@{}}
			\toprule
			Setting & Description & Motivation & Proteins & Metric \\
			\midrule
			Regression  &  
			Predict docking scores and minimize prediction error on held-out test set   & 
			Evaluation of molecular representations and predictive models   &
			PARP1, F2, KIT, ESR2, PGR  &   
			Coefficient of determination ($R^2$)  \\
			\midrule
			Virtual screening  &  
			Rank molecules according to their docking score and compute enrichment in top-$k$-ranking molecules &
			Model evaluation for hit discovery in large molecular libraries  &
			PARP1, KIT, PGR &   
			Enrichment factor (EF) \\
			\midrule
			\textit{De novo} design  &  
			Given a training dataset and a fixed budget of objective function evaluations,
			propose molecules that optimize an objective & 
			Model evaluation for hit discovery by \textit{de novo} molecular design &
			\Gape[15pt][0pt]{\makecell{F2 \\[0.3em] PPAR\{A, D, G\} \\[0.3em] JAK2, LCK}} & 
			\makecell[lc]{
			$f_\text{F2}(\ell) \coloneqq s(\ell, \text{F2}) + 10 \left(1 - \text{QED}(\ell) \right)$ \\[0.3em]
			$f_\text{PPAR}(\ell) \coloneqq \max_{t \in \text{PPAR}} s(\ell, t) + 10 \left( 1 - \text{QED}(\ell) \right)$ \\[0.3em]
			$f_\text{JAK2}(\ell) \coloneqq s(\ell, \text{JAK2}) - \min \left(s(\ell, \text{LCK}), -8.1 \right) + 10 \left(1 - \text{QED}(\ell)\right)$
			}
            \\
			\bottomrule
		\end{tabular}
		}
\end{table}

%%%%%%%%%%%%%%%%%%%%%%%%%%%%%%%%%%%%%%%%%%%%%%%%%%%%%%%%%%%%
\subsubsection{Regression}
\label{ssec:regression-benchmark-methods}
%%%%%%%%%%%%%%%%%%%%%%%%%%%%%%%%%%%%%%%%%%%%%%%%%%%%%%%%%%%%
%
\compactparagraph{Task description.}
For each target, the task is to train a regression model
to predict the docking score of a given SMILES string.
The models were trained and tested on the \dockstring{} dataset,
split into training and test sets according to the cluster labels (see Section~\ref{ssec:ligand_selection}). 
Cluster splitting is recommended because chemical datasets contain many analogous (yet unique) molecules, 
such that simple random split will likely lead to an overestimation of test performance.

\compactparagraph{Proposed benchmark.}
While all targets could be used in this benchmark,
the large number of targets in our dataset would make this benchmark expensive and difficult to interpret.
Therefore, we selected five targets from different protein families whose docking scores were deemed of high quality, 
based on enrichment analysis of experimental activity labels (Section~\ref{sssec:target_selection}).
To ensure that we included a range of difficulties, we performed an initial experiment where we regressed the docking scores of all high-quality targets.
We found that performance varied considerably depending on the target and the method employed, with coefficients of determination $R^2$ ranging between $0.2$ and $0.9$ (details in Table~\ref{tab:regression} of the \gls{si}).
Based on these results, we proposed the five following benchmark targets (with protein function and level of difficulty in brakets):
PARP1 (enzyme, easy),
F2 (protease, easy to medium),
KIT (kinase, medium),
ESR2 (nuclear receptor, hard),
and PGR (nuclear receptor, hard).

%%%%%%%%%%%%%%%%%%%%%%%%%%%%%%%%%%%%%%%%%%%%%%%%%%%%%%%%%%%%
\subsubsection{Virtual Screening}
\label{ssec:virtual_screening-methods}
%%%%%%%%%%%%%%%%%%%%%%%%%%%%%%%%%%%%%%%%%%%%%%%%%%%%%%%%%%%%

\compactparagraph{Task description.}
The goal of screening is to identify actives from a large library that is too big for detailed experimental analysis.
Screening methods attempt to solve this issue by first scoring the library and selecting a smaller subset enriched with high scores (in virtual screening, the scoring is done computationally).
Then, the subset can be studied in more detail.
A metric that is typical of screening experiments is the enrichment factor (EF), 
defined as the rate of actives in the selected subset over the rate of actives in the initial library.
Here, we propose to rank all (around 1 billion) compounds in the ZINC20 database of commercially available druglike molecules \cite{irwin2012zinc,irwin2020zinc20} and compute the enrichment factor (EF) of the top-ranking subset.
Since ground-truth labels are not available for all molecules in ZINC20,
we mark a molecule as active if its docking score is lower (i.e., better) than a certain threshold.
We chose this threshold to be the lowest $0.1$ percentile in the ZINC20 database,
which we estimated from a random sample of 100K molecules. Therefore, the maximum possible enrichment in our experiments is 1000.

\compactparagraph{Proposed benchmark.}
We trained models on the docking scores of PARP1, KIT, and PGR using all molecules in our dataset.
As in the regression benchmark, these targets were chosen to represent a range of regression difficulties.
Trained models were used to rank all the molecules in ZINC20 and select the top 5000 compounds with the lowest predicted scores.
Once the most promising molecules had been selected, we computed their actual docking scores with \dockstring{}.
Molecules were labelled as active if their actual scores were below the 0.1 percentile threshold which was 
-10.7 for KIT, -12.1 for PARP1, and -10.1 for PGR.
Finally, the enrichment factor (EF) was computed as the ratio of active molecules in the selected subset over the ratio of active molecules in ZINC20.
Note that the latter is 0.1\% by design.

Note that the virtual screening benchmark is distinct from the regression benchmark
in that it only requires \emph{ranking} compounds instead of explicitly predicting docking scores.
Further, it uses a different evaluation metric, the EF, which is uncommon in the ML literature but popular in cheminformatics.
This metric evaluates only the top molecules, whereas regression metrics depend on the predictions of all molecules.

%%%%%%%%%%%%%%%%%%%%%%%%%%%%%%%%%%%%%%%%%%%%%%%%%%%%%%%%%%%%
\subsubsection{\textit{De novo} Molecular Design}
\label{ssec:molopt-methods}
%%%%%%%%%%%%%%%%%%%%%%%%%%%%%%%%%%%%%%%%%%%%%%%%%%%%%%%%%%%%

\compactparagraph{Task description.}
The goal of \textit{de novo} design is to propose novel molecules that optimize an objective function given a certain budget. 
To be representative of real problems in drug discovery, 
this budget should be high enough to allow for significant exploration but small enough to resemble the experimental budget of a committed wet lab.

Docking scores are biased towards high molecular weight and lipophilicity. 
Therefore, optimizing docking scores alone can lead to large and hydrophobic molecules, 
as we observed in our initial experiments (Figure~\ref{fig:all-tasks-topmols}). 
These compounds are undesirable because they will suffer from poor ADMET properties and off-target effects\cite{carta2007unbiasing,hopkins2014role}.
We found that adding a druglikeness penalty based on \gls{qed} helped remedy this issue.

\compactparagraph{Proposed benchmark.}
The goal of each \textit{de novo} design task is to minimize a docking-based objective function, 
having access to the whole dataset and 5000 function evaluations. 
In the case of predictive generative models such as GP-BO, the whole dataset could be used to learn the docking score function, whereas in genetic algorithms it could be used to set the initial population.
We propose three objective functions, all of which contain a weighted \gls{qed} term to promote druglikeness.
Let $t$ be a target, $\ell$ be a ligand, and $s(\ell, t)$ be the docking score of $\ell$ against $t$.
Let $\text{QED}(\ell)$ be the \gls{qed} value of $\ell$. 

\begin{enumerate}
	\item \textbf{F2}:
	      a comparatively easy task that requires docking well to a single protein.
	      \begin{equation}
	      f_\text{F2}(\ell) = s(\ell, \text{F2}) + 10 \big(1 - \text{QED}(\ell)\big)
	      \end{equation}

	\item \textbf{Promiscuous PPAR}:
	      requires strong binding to the three PPAR nuclear receptors.
	      PPAR scores are positively correlated, so this is a task of medium difficulty.
	      ``Promiscuous'' pan-PPAR agonists are being researched as treatments against metabolic syndrome \cite{pan_ppar}.
	      If $\text{PPAR} \coloneqq \{\text{PPARA}, \text{PPARD}, \text{PPARG}\}$, then the objective function is
	      \begin{equation}
	      f_\text{PPAR}(\ell) = \max_{t\in \text{PPAR}} s(\ell, t) + 10 \big(1 - \text{QED}(\ell)\big)
	      \end{equation}

	\item \textbf{Selective JAK2}:
	      requires strong binding to JAK2 and weak binding to LCK.
	      The challenge is that, since they are both kinases, their scores are positively correlated ($\rho = 0.80$).
	      Due to their role in cell signaling and cancer, kinases are highly relevant targets but achieving
	      selectivity is notoriously difficult, and off-target effects and toxicity are common \cite{ferguson2018kinase}.
	      Our proposed objective anchors the LCK score to its median ($-8.1$):
	      \begin{equation}
	      f_\text{JAK2}(\ell) = s(\ell, \text{JAK2}) - \min \big(s(\ell, \text{LCK}), \, -8.1\!\big) + 10 \big(1 - \text{QED}(\ell)\big)
	      \end{equation}
\end{enumerate}

%%%%%%%%%%%%%%%%%%%%%%%%%%%%%%%%%%%%%%%%%%%%%%%%%%%%%%%%%%%%
\subsection{Baselines}
\label{ssec:baseline_methods}
%%%%%%%%%%%%%%%%%%%%%%%%%%%%%%%%%%%%%%%%%%%%%%%%%%%%%%%%%%%%

We tested a variety of classical and more modern algorithms to assess the difficulty of the \dockstring{} benchmarks tasks. 
Training and testing datasets and procedures followed each tasks' specifications as described in Section~\ref{sec:benchmarks-methods}.
Additional details are given in the \gls{si}.

%%%%%%%%%%%%%%%%%%%%%%%%%%%%%%%%%%%%%%%%%%%%%%%%%%%%%%%%%%%%
\subsubsection{Regression and virtual screening}
%%%%%%%%%%%%%%%%%%%%%%%%%%%%%%%%%%%%%%%%%%%%%%%%%%%%%%%%%%%%

\paragraph{\texttt{scikit-learn} algorithms.}
Ridge and lasso regression were implemented with \texttt{scikit-learn}.
XGBoost was implemented with the XGBoost library \cite{Chen:2016:XST:2939672.2939785}
using the \texttt{scikit-learn} API.
For all these methods, hyperparameter selection was done with random search over 20 configurations,
evaluating each configuration using a 5-fold cross-validation score
(implemented via \texttt{scikit-learn}'s \texttt{RandomizedSearchCV} function).

\paragraph{Gaussian processes.}
All \gls{gp} algorithms used the Tanimoto kernel \cite{tanimoto1958elementary} with fingerprint features.
Due to the cubic scaling of GP regression,
the exact \gls{gp} was trained on 10K randomly chosen data points.
In comparison, the sparse \gls{gp} used 10K randomly chosen training points as the inducing variables,
but was trained on the whole dataset.
Hyperparameters were chosen by maximizing the log-marginal likelihood on the training set.
All \glspl{gp} were implemented with PyTorch \cite{pytorch-paper} and GPyTorch \cite{gardner2018gpytorch}.

\paragraph{Graph neural networks.}
The DeepChem library's implementation of the Attentive FP and MPNN was used \cite{Ramsundar-et-al-2019}.
Both models were trained with default parameters from the DeepChem library for 10 epochs.
Preliminary experiments with a third method from DeepChem,
the Graph Attention Network \cite{velivckovic2018graph},
were performed but the model frequently overfitted to the training data;
we decided to omit it rather than tune the hyperparameters for this model.

%%%%%%%%%%%%%%%%%%%%%%%%%%%%%%%%%%%%%%%%%%%%%%%%%%%%%%%%%%%%
\subsubsection{De novo design}
%%%%%%%%%%%%%%%%%%%%%%%%%%%%%%%%%%%%%%%%%%%%%%%%%%%%%%%%%%%%

\paragraph{Graph genetic algorithm.}
The implementation from the GuacaMol baselines \cite{brown2019guacamol} was used\cite{guacamol_baselines_url}.
The population size was set to 250,
the offspring size to 25,
and the mutation rate to 0.01.
The population size was chosen based on some preliminary experiments
with the GuacaMol dataset,
and the offspring size was arbitrarily chosen to be 25
to allow for 200 generations to occur.
The value of the mutation rate was the default used in the GuacaMol implementation.

\paragraph{SELFIES genetic algorithm.}
The implementation of the SELFIES genetic algorithm was taken from the GitHub
repository of Nigam et al~\cite{nigam2021beyond}.
It is a simple genetic algorithm which randomly inserts, deletes, 
or modifies a single token of a SELFIES string \cite{krenn2020self}.
The algorithm was not tuned and represents the minimum level of performance that can be expected from any reasonable genetic algorithm.
The offspring and population size hyperparameters were the same as for the graph genetic algorithm.

\paragraph{Bayesian optimization.}
The \gls{gp} implementation is identical to the exact \gls{gp} implementation from Section~\ref{ssec:regression-benchmark} using the Tanimoto kernel \cite{tanimoto1958elementary,ralaivola2005graph}.
As it is computationally infeasible to train a \gls{gp} on the entire dataset,
the 2000 training points with the smallest objective score
and 3000 random points were selected from the dataset for training.
Kernel hyperparameters were chosen by maximizing the log marginal likelihood
on this training set.
At each iteration, a batch of five new molecules was selected by maximizing either
the upper confidence bound acquisition function \cite{srinivas2010gaussian} with $\beta=10$ (i.e.,\@ $\mu+10\sigma$)
or the expected improvement acquisition function \cite{jones1998efficient}.
$\beta$ was chosen based on the \gls{gp} hyperparameters from Section~\ref{ssec:regression-benchmark}
and a small amount of preliminary
experiments on the GuacaMol benchmarks to encourage a combination of exploration and exploitation,
but was not tuned once experiments on the docking objectives were started.
Optimization was done using the graph genetic algorithm as described above,
with an offspring size of 1000 and 25 generations.
The batch was then scored and the \gls{gp} retrained using the new scores, with the hyperparameters remaining unchanged.
This was repeated until the objective function evaluation budget was reached.

%%%%%%%%%%%%%%%%%%%%%%%%%%%%%%%%%%%%%%%%%%%%%%%%%%%%%%%%%%%%
\section{Results}
%%%%%%%%%%%%%%%%%%%%%%%%%%%%%%%%%%%%%%%%%%%%%%%%%%%%%%%%%%%%

This section introduces the three components of the \dockstring{} bundle: 
a user-friendly molecular docking package,
an extensive dataset,
and a set of challenging benchmark tasks. The package is available at \githuburl{}, while the dataset and code for benchmark baselines are available at \figshareurl{}. All components are released under the Apache 2.0 license.

%%%%%%%%%%%%%%%%%%%%%%%%%%%%%%%%%%%%%%%%%%%%%%%%%%%%%%%%%%%%
\subsection{Molecular Docking Package}
\label{sec:wrapper}
%%%%%%%%%%%%%%%%%%%%%%%%%%%%%%%%%%%%%%%%%%%%%%%%%%%%%%%%%%%%

\begin{figure}[ht]
	\centering
	\includegraphics[width=3.33in]{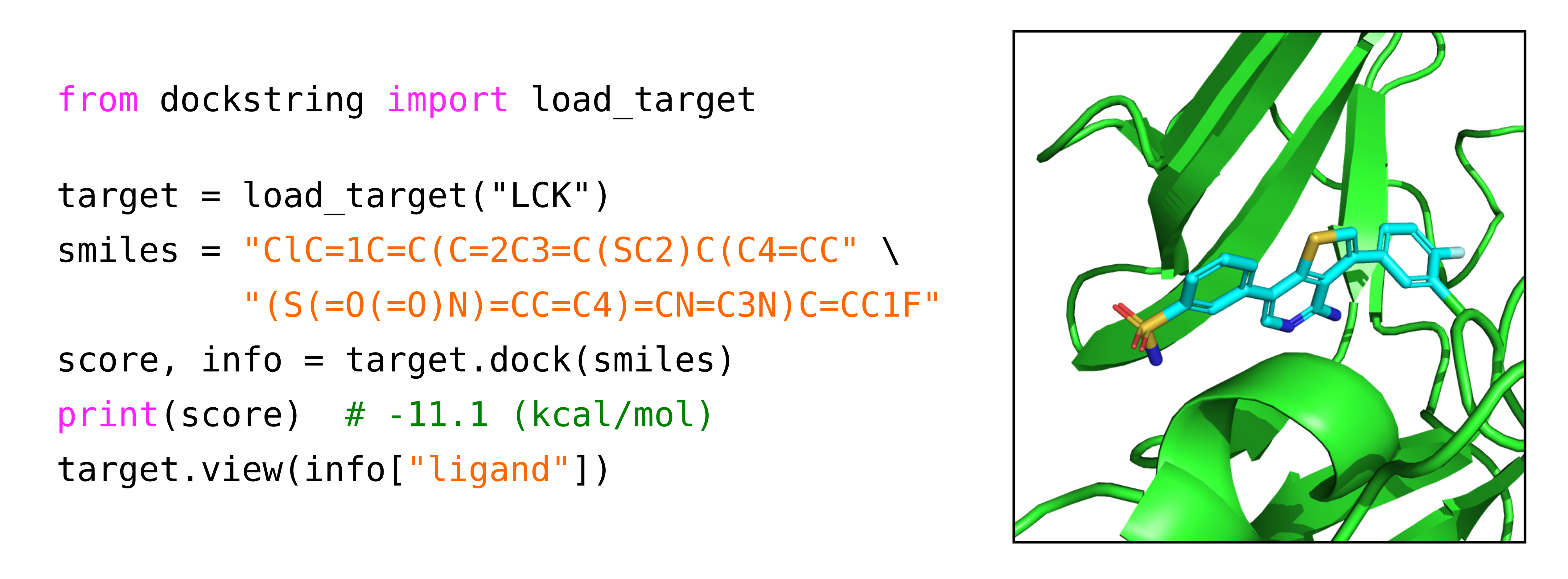}
	\caption{
		\dockstring{} provides a simple API for docking and visualization.
		User-defined targets and custom pH's can be specified if required.
		\textbf{left}: code example for docking.
		\textbf{right}: visualization of the docking pose in the active site of the target LCK.
	}
	\label{fig:api}
\end{figure}

We developed a Python package that interfaces with AutoDock Vina to allow the computation of docking scores in just a few lines of code.
The user only needs to provide the name of a target protein and the SMILES string of a ligand molecule (Figure~\ref{fig:api}, left).
The target name can be chosen from a list of 58 targets (Table~\ref{tab:all-targets}) that have been prepared as explained in Section~\ref{ssec:target_prep}.
Ligands are prepared automatically by the \dockstring{} package as explained in Section~\ref{ssec:ligand_prep}.
\dockstring{} returns up to nine docking poses with their corresponding docking scores,
which can be used in downstream tasks such as bioactivity prediction and visualization (Figure~\ref{fig:api}, right). Note that in subsequent experiments in this work, we always use the lowest (i.e. best) docking score.
For most targets, computing a score with eight CPUs takes around 15s (Table~\ref{tab:docking-times} of the \gls{si}).
We also note that, by default, our docking wrapper carefully controls all sources of randomness in the docking
procedure so that the output is deterministic (see Section~\ref{ssec:docking}).
Finally, the target, the search box, and all poses can be visualized with the PyMOL software package.

% \begin{figure}[ht]
% 	\centering%
% 	\begin{minipage}{3.33in}
% 		\begin{lstlisting}[language=Python,style=mystyle,label={lst:api}]
% from dockstring import load_target
% 
% target = load_target("LCK")
% smiles = "ClC=1C=C(C=2C3=C(SC2)C(C4=CC" \
%     "(S(=O)(=O)N)=CC=C4)=CN=C3N)C=CC1F"
% score, info = target.dock(smiles)
% 
% print(score)  # -11.1 (kcal/mol)
% target.view(info["ligand"])
%     \end{lstlisting}
% 	\end{minipage}
% 	\begin{minipage}{0.35\textwidth}
% 		\centering
% 		\includegraphics[width=0.60\textwidth]{figures/api.png}
% 	\end{minipage}
% 	\caption{
% 		\dockstring{} provides a simple API for docking and visualization.
% 		User-defined targets and custom pH's can be specified if required.
% 		\textbf{left}: code example for docking.
% 		\textbf{right}: visualization of the docking pose in the active site of the target LCK.
% 	}
% 	\label{fig:api}
% \end{figure}

%%%%%%%%%%%%%%%%%%%%%%%%%%%%%%%%%%%%%%%%%%%%%%%%%%%%%%%%%%%%
\subsection{Dataset}
\label{sec:dataset}
%%%%%%%%%%%%%%%%%%%%%%%%%%%%%%%%%%%%%%%%%%%%%%%%%%%%%%%%%%%%

Molecular docking is applicable in areas such as regression, molecular optimization, virtual screening, transfer learning, multi-task learning, and representation learning.
Since most of these settings require an initial training dataset, 
we provide a set of more than 15 million scores for a diverse and highly-curated set of more than $260,000$ ligands 
docked against 58 targets.
This dataset required more than 500K CPU hours to compute (see Section~\ref{sec:comp_details} of the \gls{si} for computational details).
The target and ligand selection process are detailed below.

%%%%%%%%%%%%%%%%%%%%%%%%%%%%%%%%%%%%%%%%%%%%%%%%%%%%%%%%%%%%
\subsubsection{Target Selection}
\label{sssec:target_selection}
%%%%%%%%%%%%%%%%%%%%%%%%%%%%%%%%%%%%%%%%%%%%%%%%%%%%%%%%%%%%

Our dataset comprises 58 targets covering a variety of protein functions:
kinases (22),
enzymes (12),
nuclear receptors (9),
proteases (7),
G-protein coupled receptors (5),
cytochromes (2),
and chaperone (1).
For details, see Table~\ref{tab:all-targets}.
We have identified a subset of 24 targets whose docking scores are of relatively high quality based on their ability to achieve enrichment of experimental active labels (details are given in Section~\ref{ssec:docking_scores}).
These high-quality targets are involved in a range of diseases
and are thus considered of great interest in drug discovery (examples can be seen in Table~\ref{tab:drugs-approved} of the \gls{si}).

\begin{table}[ht]
	\centering
	\caption{
		Genes of targets in the \dockstring{} dataset grouped by function and quality of docking scores (***: best, *: worst).
	}
	\label{tab:all-targets}
	\resizebox{\columnwidth}{!}{%
		\begin{tabular}{@{}lll@{}}
			\toprule
			Group            & Quality & Gene                                                           \\
			\midrule
			Kinase           & ***     & IGF1R, JAK2, KIT, LCK, MAPK14, MAPKAPK2, MET, PTK2, PTPN1, SRC \\
			                 & **      & ABL1, AKT1, AKT2, CDK2, CSF1R, EGFR, KDR, MAPK1, FGFR1, ROCK1  \\
			                 & *       & MAP2K1, PLK1                                                   \\
			Enzyme           & ***     & HSD11B1, PARP1, PDE5A, PTGS2                                   \\
			                 & **      & ACHE, MAOB                                                     \\
			                 & *       & CA2, GBA, HMGCR, NOS1, REN, DHFR                               \\
			Nuclear Receptor & ***     & ESR1, ESR2, NR3C1, PGR, PPARA, PPARD, PPARG                    \\
			                 & **      & AR                                                             \\
			                 & *       & THRB                                                           \\
			Protease         & ***     & ADAM17, F10, F2                                                \\
			                 & **      & BACE1, CASP3,  MMP13                                           \\
			                 & *       & DPP4                                                           \\
			GPCR             & **      & ADRB1, ADRB2, DRD2, DRD3                                       \\
			                 & *       & ADORA2A                                                        \\
			Cytochrome       & **      & CYP2C9, CYP3A4                                                 \\
			Chaperone        & *       & HSP90AA1                                                       \\
			\bottomrule
		\end{tabular}
	}
\end{table}

%%%%%%%%%%%%%%%%%%%%%%%%%%%%%%%%%%%%%%%%%%%%%%%%%%%%%%%%%%%%
\subsubsection{Ligand Selection and Clustering}
\label{ssec:ligand_selection}
%%%%%%%%%%%%%%%%%%%%%%%%%%%%%%%%%%%%%%%%%%%%%%%%%%%%%%%%%%%%

\begin{figure}[ht]
	\centering
	\includegraphics{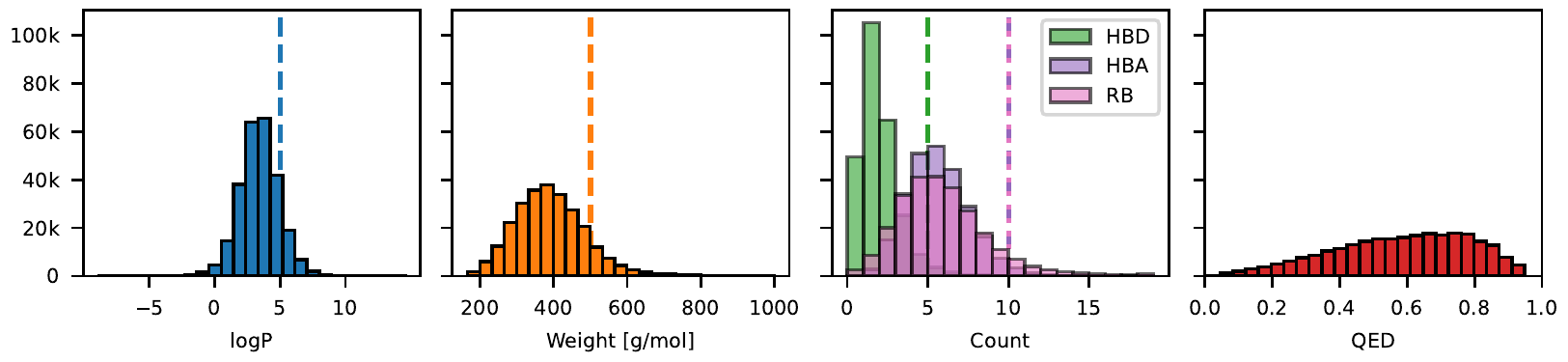}
	\caption{
		Distribution of molecular properties of ligands in the \dockstring{} dataset.
		Most ligands in our dataset fulfill ``Lipinski's rules of five'' \cite{Lipinski2001Experimental} (vertical dashed lines) for the properties depicted (logP, molecular weight, number of hydrogen bond donors [HBD], hydrogen bond acceptors [HBA] and rotatable bonds [RB]).
		In addition, the QED distribution is left-skewed and peaks at 0.75,
		further suggesting that most ligands in our dataset are druglike.
	}
	\label{fig:dataset}
\end{figure}

ExCAPE is a large database that aggregates results from a variety of assays in PubChem and ChEMBL, many of them from real screening experiments for hit discovery.
Furthermore, it sets explicit filters for physicochemical properties such as molecular weight and number of heavy atoms to further promote druglikeness.
For these reasons, ExCAPE is generally regarded as diverse and druglike.
Indeed, we found that most ligands fulfill Lipinski's rules \cite{Lipinski2001Experimental} and feature favorable \gls{qed} profiles (Figure~\ref{fig:dataset}).

\begin{figure}[ht]
	\centering
	\includegraphics{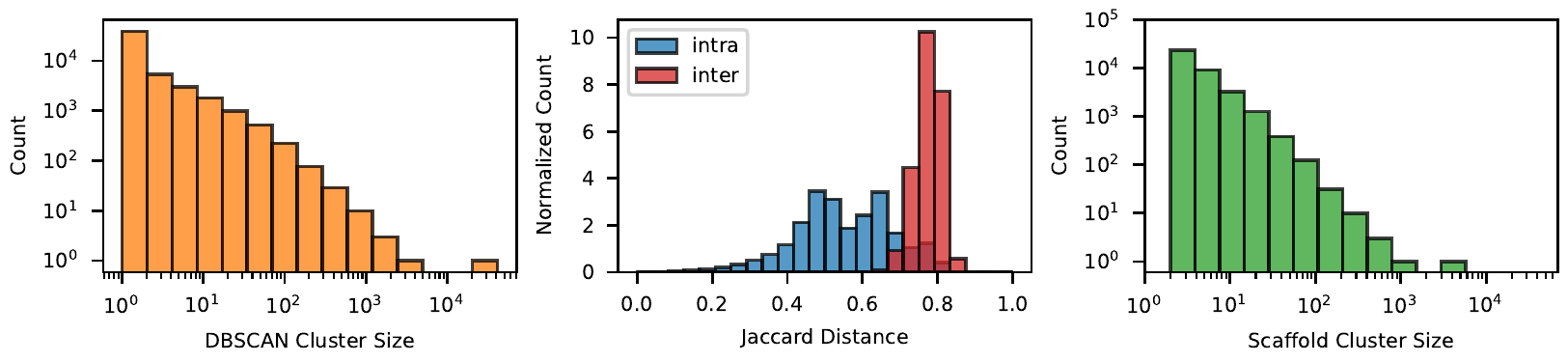}
	\caption{
		Cluster analysis of \dockstring{} dataset.
		\textbf{(left)}: distribution of clusters grouped by the DBSCAN algorithm using the Tanimoto distance.
		\textbf{(middle)}: normalized count of Jaccard distances between molecules within the same cluster (blue) and between different ones (red).
		\textbf{(right)}: distribution of clusters grouped by scaffold.
		Here, only molecules from the second and third largest clusters are considered.
	}
	\label{fig:clusters}
\end{figure}

% Clustering
We performed cluster analyses with two different techniques: 
DBSCAN (Density-Based Spatial Clustering of Applications with Noise), a data-type agnostic clustering algorithm,
and Bemis-Murcko scaffold decomposition, which is especially designed for molecules.
Given a cluster and a query point, DBSCAN assigns a point to the cluster if it is within the $\varepsilon$-neighborhood of one of its core points 
(where a core point is one that has a minimum number of neighbors from the same cluster)\cite{ester1996a}.
DBSCAN found 52K clusters, where the biggest one covered over 15\% of the dataset 
and 31K clusters contained only a single molecule (Figure~\ref{fig:clusters}, left).
The Jaccard distance within the same cluster was significantly smaller than the distance between different clusters, with little overlap of the two (Figure~\ref{fig:clusters}, middle).

Bemis-Murcko decomposition is rooted in the concept of molecular scaffolds \cite{bemis1996the}.
A scaffold is defined as the union of the ring systems in a molecule plus the linker atoms between them.
Thus, there are many possible molecules with the same scaffold that differ only in their side chains and atom types.
Molecules with the same scaffold are structurally similar and are expected to have similar properties.
We found that our ligand set contains 102K Bemis-Murcko scaffolds.
They showed a similar distribution to DBSCAN clusters, with the most popular scaffold standing out from the rest and 64K single-molecule scaffold clusters (Figure~\ref{fig:clusters}, right).
Overall, these results confirm that our ligand set is diverse.

%%%%%%%%%%%%%%%%%%%%%%%%%%%%%%%%%%%%%%%%%%%%%%%%%%%%%%%%%%%%
\subsubsection{Docking Scores}
\label{ssec:docking_scores}
%%%%%%%%%%%%%%%%%%%%%%%%%%%%%%%%%%%%%%%%%%%%%%%%%%%%%%%%%%%%

We computed docking scores for every target-ligand pair in our dataset, resulting in more than 15M data points (see Section~\ref{sec:comp_details} of the \gls{si} for computational details).
To our knowledge, this is the first dataset that computes the full score matrix of a large ligand set against a high number of protein targets,
making it ideal for the design of meaningful benchmark tasks in settings such as multiobjective optimization and transfer learning.

\begin{figure}[h]
	\centering
	\includegraphics{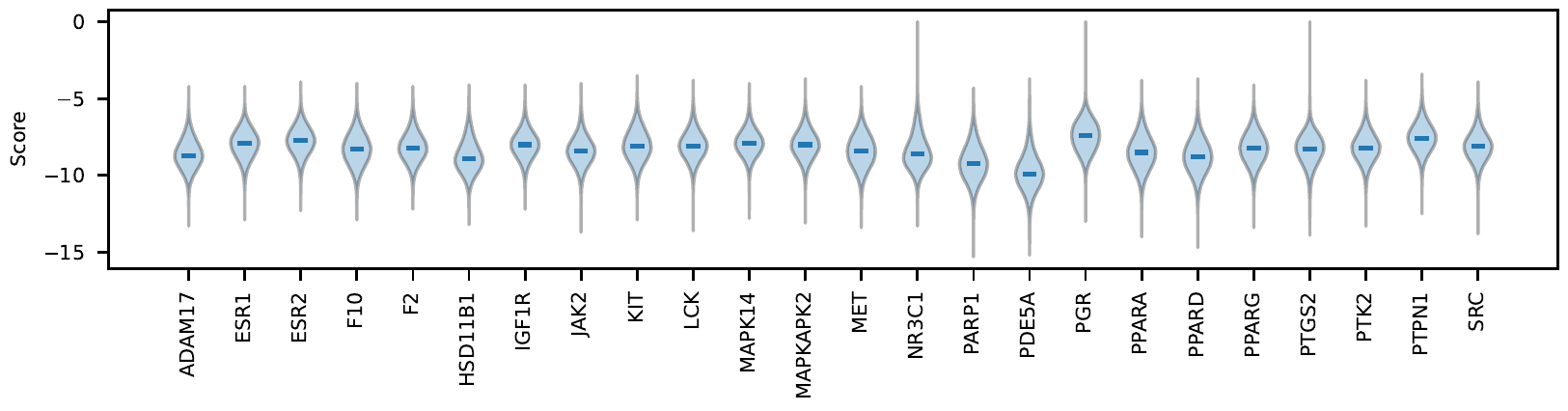}
	\caption{
		Distribution over docking scores (in kcal/mol) for a subset of high-quality targets in the \dockstring{} dataset in alphabetical order.
		The tails of each violin plot represent the minimum and maximum docking score for each target.
		The blue vertical bars indicate the median.
		For this plot, docking scores greater than zero were set to zero.
	}
	\label{fig:violin}
\end{figure}

\begin{figure}[ht]
	\centering
	\includegraphics{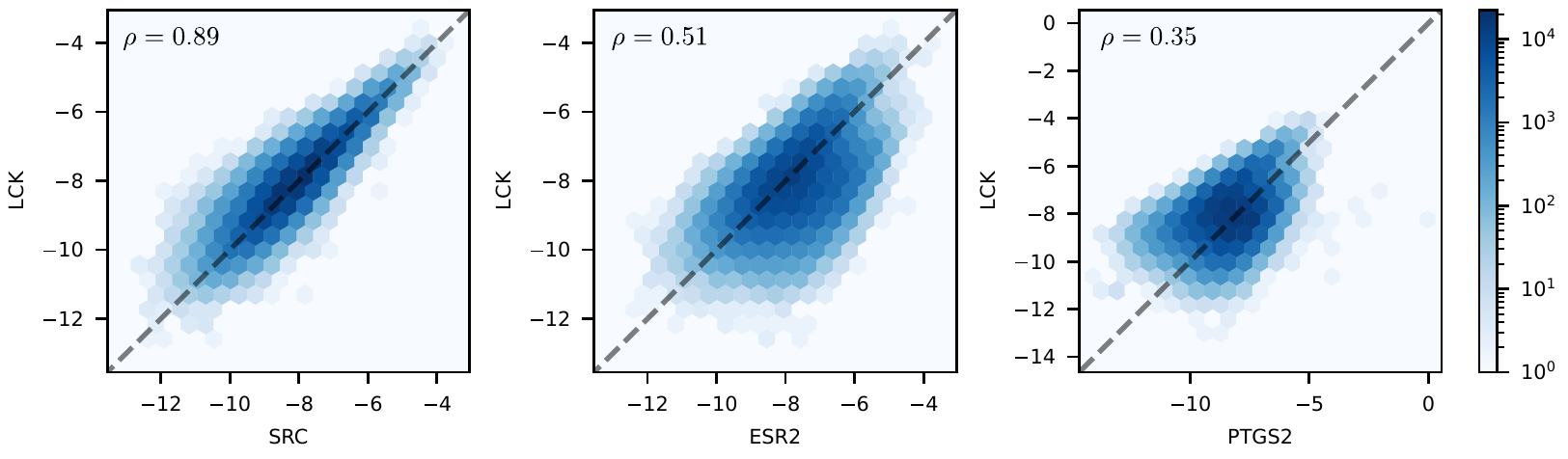}
	\caption{
		Correlations of docking scores (in kcal/mol) between the kinase LCK and three other targets from the \dockstring{} dataset:
		SRC, a target from the same family \textbf{(left)},
		ESR2, a nuclear receptor \textbf{(middle)},
		and PTGS2, a cyclooxygenase \textbf{(right)}.
		Unlike target independent molecular properties (e.g., logP and \gls{qed}),
		docking scores can vary significantly between targets depending on their structural similarity.
	}
	\label{fig:score_correlation}
\end{figure}

% Results
We found that the docking scores were similarly distributed for most proteins, ranging between $-4$ and $-13$, as can be seen in Figure~\ref{fig:violin} (note that in the original AutoDock Vina publication\cite{Trott2010AutoDock} scores are reported in kcal/mol, but for our purposes scores can be treated as a unitless quantity).
Docking scores can be interpreted as the binding free energy, so more negative scores suggest stronger binding.
We also found that high-quality targets that were functionally related or were homologs
(i.e., proteins with high sequence similarity such as ESR1 and ESR2)
exhibited high correlation, whereas unrelated targets tended to show medium or poor correlation (Figure~\ref{fig:score_correlation}).
This supports the claim that \dockstring{} scores are biologically meaningful.

\begin{figure}[h]
	\centering
	\includegraphics{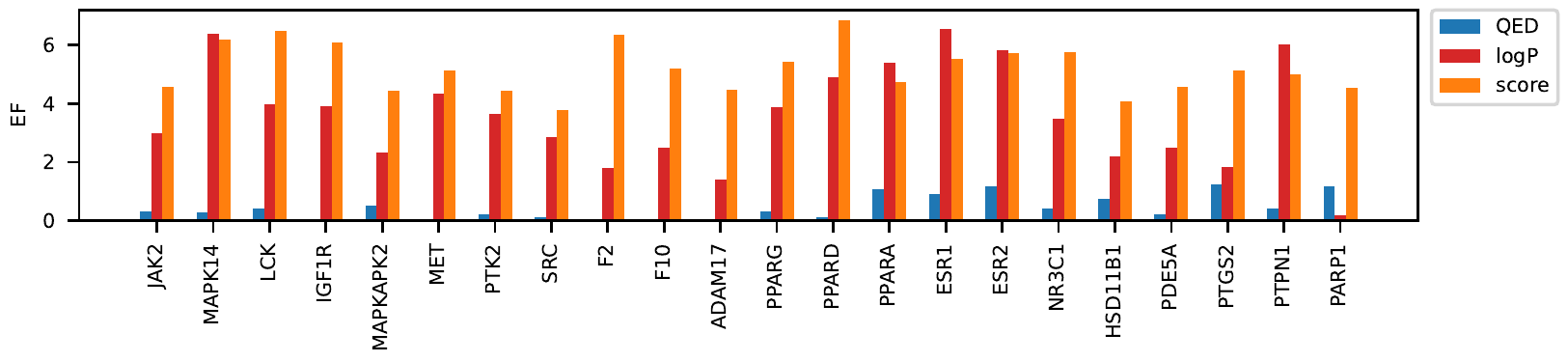}
	\caption{
		Enrichment factor (EF) of the docking score (orange) and two target-independent molecular properties, \gls{qed} (blue) and logP (red),
		for the high-quality targets in the \dockstring{} dataset in alphabetical order.
		For most targets, docking scores yielded higher EF than that of the logP or \gls{qed}.
	}
	\label{fig:quality_metrics}
\end{figure}

We assessed the quality of each target's docking scores based on their enrichment factor (EF),
using experimental activity labels from ExCAPE as reference (Figure~\ref{fig:quality_metrics}).
Such assessment was necessary because docking is known to perform differently on different proteins, and the optimal docking workflow may vary from one protein to another \cite{su2019comparative}.
We found that docking scores achieved the highest enrichment overall, although they were surpassed by a small difference by logP in a few targets. This result can be explained because greasy molecules bind non-specifically to many targets with hydrophobic pockets.
However, since this kind of binding is not selective, it may reduce efficacy and increase the risk of toxicity.
Therefore, molecules with high logP are usually discarded in drug discovery projects \cite{johnson2018lipophilic}. Finally, \gls{qed} achieved very low to no enrichment.
Overall, our results indicate that our preparation and docking protocols are effective and yield meaningful docking scores.

%%%%%%%%%%%%%%%%%%%%%%%%%%%%%%%%%%%%%%%%%%%%%%%%%%%%%%%%%%%%
\subsubsection{Docking Poses}
\label{ssec:docking_poses}
%%%%%%%%%%%%%%%%%%%%%%%%%%%%%%%%%%%%%%%%%%%%%%%%%%%%%%%%%%%%

A typical docking simulation results in two outputs: docking poses, which are conformations of the ligand in the binding pocket,
and their corresponding docking scores, which quantify the strength of the ligand-target interaction.
Scores are convenient for ranking compounds in virtual screening workflows. 
However, they are an approximate heuristic and provide little insight into protein-ligand interactions. 
By contrast, poses are more interpretable and can help discriminate false positives.
Finally, poses can be used as input to ML algorithms that exploit 3D structure information.
An example of such models are ML-based scoring functions which produce docking scores from docking poses,
which have attracted considerable interest in recent years \cite{li2021machine-learning}.
For these reasons, each docking score in our dataset is released together with its corresponding docking pose, adding up to more than 15M conformations.
To our knowledge, the \dockstring{} dataset is the first to include this type of information.

%%%%%%%%%%%%%%%%%%%%%%%%%%%%%%%%%%%%%%%%%%%%%%%%%%%%%%%%%%%%
\subsection{Benchmarks}
\label{sec:benchmarks}
%%%%%%%%%%%%%%%%%%%%%%%%%%%%%%%%%%%%%%%%%%%%%%%%%%%%%%%%%%%%

%%%%%%%%%%%%%%%%%%%%%%%%%%%%%%%%%%%%%%%%%%%%%%%%%%%%%%%%%%%%
\subsubsection{Regression}
\label{ssec:regression-benchmark}
%%%%%%%%%%%%%%%%%%%%%%%%%%%%%%%%%%%%%%%%%%%%%%%%%%%%%%%%%%%%

A variety of classical regression algorithms were trained on 1024-dimensional binary Morgan fingerprints \cite{rogers2010extended} with a radius of two:
ridge and lasso regression \cite{Hastie2009Elements},
gradient-boosted decision trees (XGBoost) \cite{friedman2001greedy},
exact \glspl{gp} \cite{williams2006gaussian},
and sparse \glspl{gp} \cite{titsias2009variational}.
In addition, two newer algorithms leveraging graph neural networks were also employed, namely,
MPNN \cite{gilmer2017neural}
and Attentive FP \cite{xiong2019pushing}.

The regression performance of the baselines on the five benchmark targets is shown in Table~\ref{tab:regression-shortlist}.
Performance on predicting logP and \gls{qed} are also shown to help gauge the relative difficulty of the proposed tasks.
First, note that classical methods are handily outperformed by deep learning methods.
The worst ranking methods are ridge and lasso regression, which are linear models and yield coefficients of determination $R^2$ ranging between $0.242$ and $0.706$. 
In contrast, the best ranking model is Attentive FP, a graph deep neural network, 
with coefficients ranging between $0.627$ and $0.910$ and beating every other method by a significant margin.
Second, note that some targets seem to be more difficult than others.
The easiest target is PARP1,
whereas the most challenging target is PGR.
This is in contrast with logP and \gls{qed},
where the graph ML methods achieve perfect or near-perfect performance.
This strongly supports the use of docking scores instead of logP and \gls{qed} to benchmark high-performing models.

\begin{table}[h]
	\centering
	\caption{
		Regression performance for select tasks
		(full results are in Tables~\ref{tab:regression} and \ref{tab:regression-full}).
		Numbers represent the mean coefficient of determination ($R^2$ score) averaged over three runs (highest is better).
		The best score in each row is in \textbf{bold}.
		The average rank includes only the \dockstring{} targets (excluding logP and \gls{qed}).
	}
	\label{tab:regression-shortlist}
	\resizebox{\columnwidth}{!}{%
		\begin{tabular}{@{}lccccccc@{}}
			\toprule
			Target                & Ridge & Lasso & XGBoost & GP (exact) & GP (sparse) & MPNN  & Attentive FP   \\
			\midrule
			logP                  & 0.640 & 0.640 & 0.734   & 0.707      & 0.716       & 0.953 & \textbf{1.000} \\
			QED                   & 0.519 & 0.483 & 0.660   & 0.640      & 0.598       & 0.901 & \textbf{0.981} \\
			\midrule
			ESR2                  & 0.421 & 0.416 & 0.497   & 0.441      & 0.508       & 0.506 & \textbf{0.627} \\
			F2                    & 0.672 & 0.663 & 0.688   & 0.705      & 0.744       & 0.798 & \textbf{0.880} \\
			KIT                   & 0.604 & 0.594 & 0.674   & 0.637      & 0.684       & 0.755 & \textbf{0.806} \\
			PARP1                 & 0.706 & 0.700 & 0.723   & 0.743      & 0.772       & 0.815 & \textbf{0.910} \\
			PGR                   & 0.242 & 0.245 & 0.345   & 0.291      & 0.387       & 0.324 & \textbf{0.678} \\
			\midrule
			\textbf{Average Rank} & 6.04  & 6.96  & 4.17    & 4.71       & 2.88        & 2.25  & \textbf{1.00}  \\
			\bottomrule
		\end{tabular}
	}
\end{table}

%%%%%%%%%%%%%%%%%%%%%%%%%%%%%%%%%%%%%%%%%%%%%%%%%%%%%%%%%%%%
\subsubsection{Virtual Screening}
\label{ssec:virtual_screening}
%%%%%%%%%%%%%%%%%%%%%%%%%%%%%%%%%%%%%%%%%%%%%%%%%%%%%%%%%%%%

The Attentive FP and ridge regression methods from Section~\ref{ssec:regression-benchmark} were selected for virtual screening.
The former was chosen for its high regression scores,
while the latter was selected based on its simplicity and low computational cost.
Implementation details were the same as for the methods detailed in Section~\ref{ssec:regression-benchmark}.

Attentive FP always had a higher EF than ridge regression (Table~\ref{tab:virtual-screening}),
mirroring its superior performance in the regression baseline.
KIT seems an easier screening target than PARP1,
even though PARP1 was the easiest regression target.
This suggests that in-distribution prediction difficulty is different than out-of-distribution prediction difficulty for the same target, highlighting the usefulness of docking engines such as the \dockstring{} Python package for out-of-distribution and prospective validation.

\begin{table}[h]
	\centering
	\caption{
		Enrichment factors (EF) for virtual screening tasks (higher is better).
		For each target, a threshold score is given below which a ligand is considered active.
		Highest possible EF for the chosen thresholds is 1000.
	}
	\label{tab:virtual-screening}
	\begin{tabular}{@{}lc|cc@{}}
		\toprule
		Target & Threshold Score & Ridge & Attentive FP   \\
		\midrule
		KIT    & -10.7           & 451.6 & \textbf{766.5} \\
		PARP1  & -12.1           & 325.9 & \textbf{472.2} \\
		PGR    & -10.1           & 120.5 & \textbf{461.3} \\
		\bottomrule
	\end{tabular}
\end{table}

%%%%%%%%%%%%%%%%%%%%%%%%%%%%%%%%%%%%%%%%%%%%%%%%%%%%%%%%%%%%
\subsubsection{\textit{De novo} Molecular Design}
\label{ssec:molopt}
%%%%%%%%%%%%%%%%%%%%%%%%%%%%%%%%%%%%%%%%%%%%%%%%%%%%%%%%%%%%

With our novel \textit{de novo} design tasks,
we compared two genetic algorithms (GAs),
a simple GA based on SELFIES \cite{krenn2020self}
and the graph GA by Jensen \cite{jensen2019a},
with Gaussian process Bayesian optimization (GP-BO) approaches
using the upper confidence bound (UCB) and expected improvement (EI) acquisition functions 
(for details, see the Section~\ref{ssec:baseline_methods}).
We also included a random baseline which randomly selected molecules from the ZINC20 dataset.

\begin{figure}[h!]
	\centering
	\includegraphics[width=2.7in]{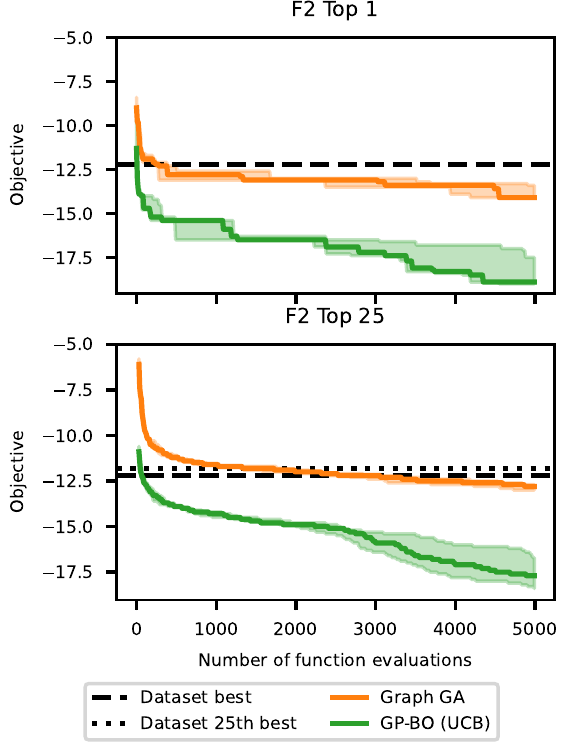}
	\caption{
		Results for baseline algorithms on the F2 \textit{de novo} molecular design task \emph{without} the QED penalty.
		The objective value of the best molecule found so far is shown as a function of the number of objective function calls.
		The solid lines indicate the median and the shaded area the minimum and maximum over three runs.
		The black dashed line indicates the best value in the \dockstring{} dataset.
	}
	\label{fig:f2-top-no-pen}
\end{figure}

\dockstring{} introduces three \textit{de novo} benchmark tasks: 
optimization of F2 docking scores (F2), 
joint optimization of PPAR nuclear receptors (Promiscuous PPAR), 
and adversarial optimization of JAK2 against LCK (Selective JAK2). 
Initially, we defined naive versions of these tasks that did not include a penalty to enforce druglikeness.
These tasks was easily solved with most methods quickly finding molecules that were far better than the best in the training set in just tens of iterations (Figure \ref{fig:f2-top-no-pen}).

\begin{figure}[h]
	\center
	\centering
	\includegraphics[width=0.95\textwidth]{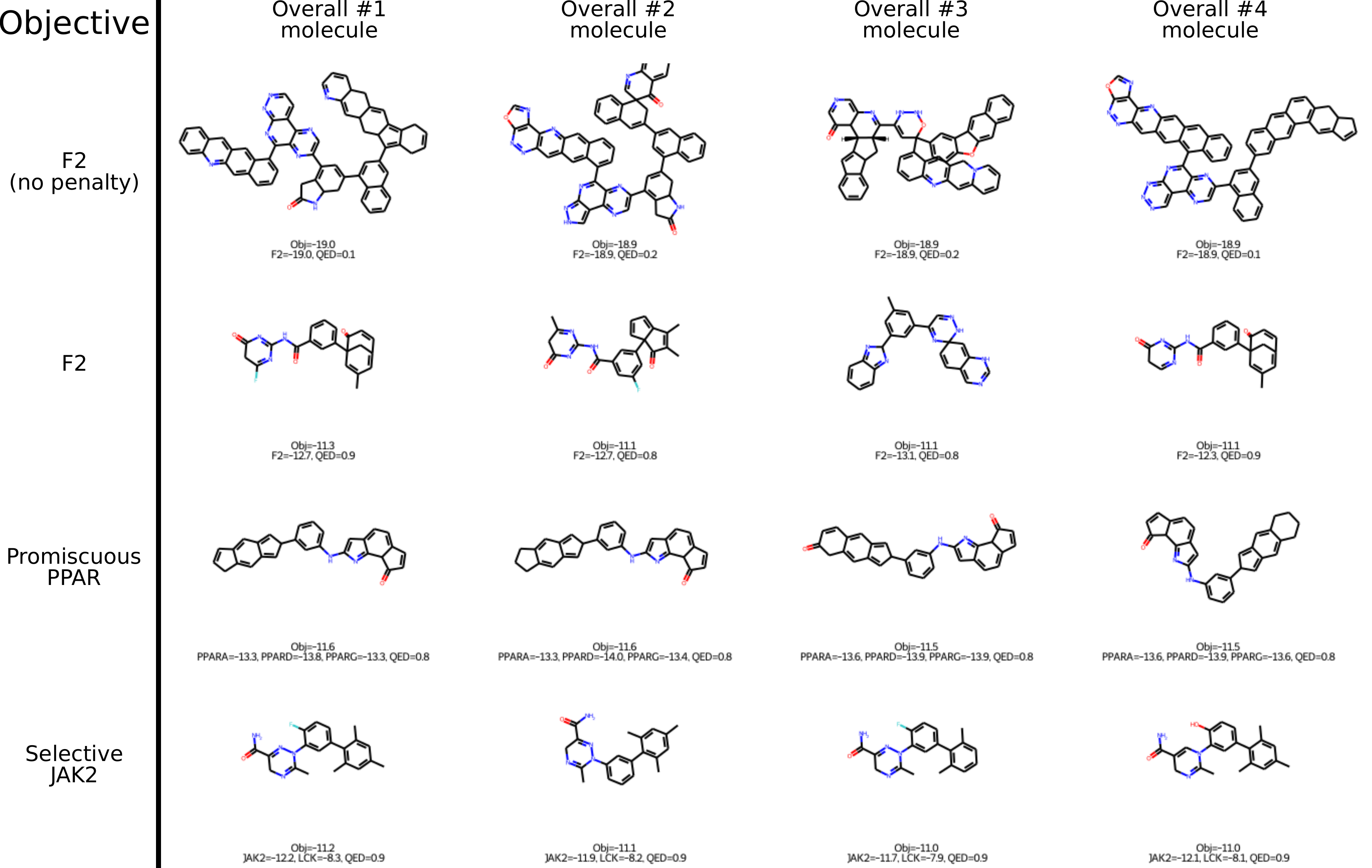}
	\caption{
		Top four molecules for F2 (no penalty), F2, Promiscuous PPAR, and Selektive JAK2.
		%\gncs{Are we keeping this Figure?}
	}
	\label{fig:all-tasks-topmols}
\end{figure}

Optimization of the naive objectives yielded molecules that were large, lipophilic and highly undruglike as per Lipinski rules and \gls{qed} (Figure~\ref{fig:all-tasks-topmols}, first row). 
This could be explained by the inherent biases of docking algorithms.
On the one hand, docking tends to give high scores to large molecules, since they can potentially establish a larger amount of interactions with the target and most scoring functions are additive.
Therefore, large molecules with high docking scores are often false positives and must be avoided \cite{carta2007unbiasing}. 
On the other hand, hydrophobic molecules bind non-specifically to many proteins with hydrophobic regions in their binding pockets, 
which can lead to off-target effects, toxicity and decreased efficiency.
Therefore, highly hydrophobic molecules are also undesirable\cite{hopkins2014role}. 

To make the tasks more challenging and enforce druglikeness explicitly, we added a \gls{qed} penalty to each of the naive tasks. 
The functional form chosen was $+ 10\left(1-\text{QED}(\ell)\right)$.
Since \gls{qed} ranges between $0$ and $1$, this penalty will be 0 at minimum and 10 at maximum, which covers approximately the same numeric value of docking scores. The full objective functions can be found in Section \ref{ssec:molopt-methods}.

\begin{figure}[h!]
	\includegraphics{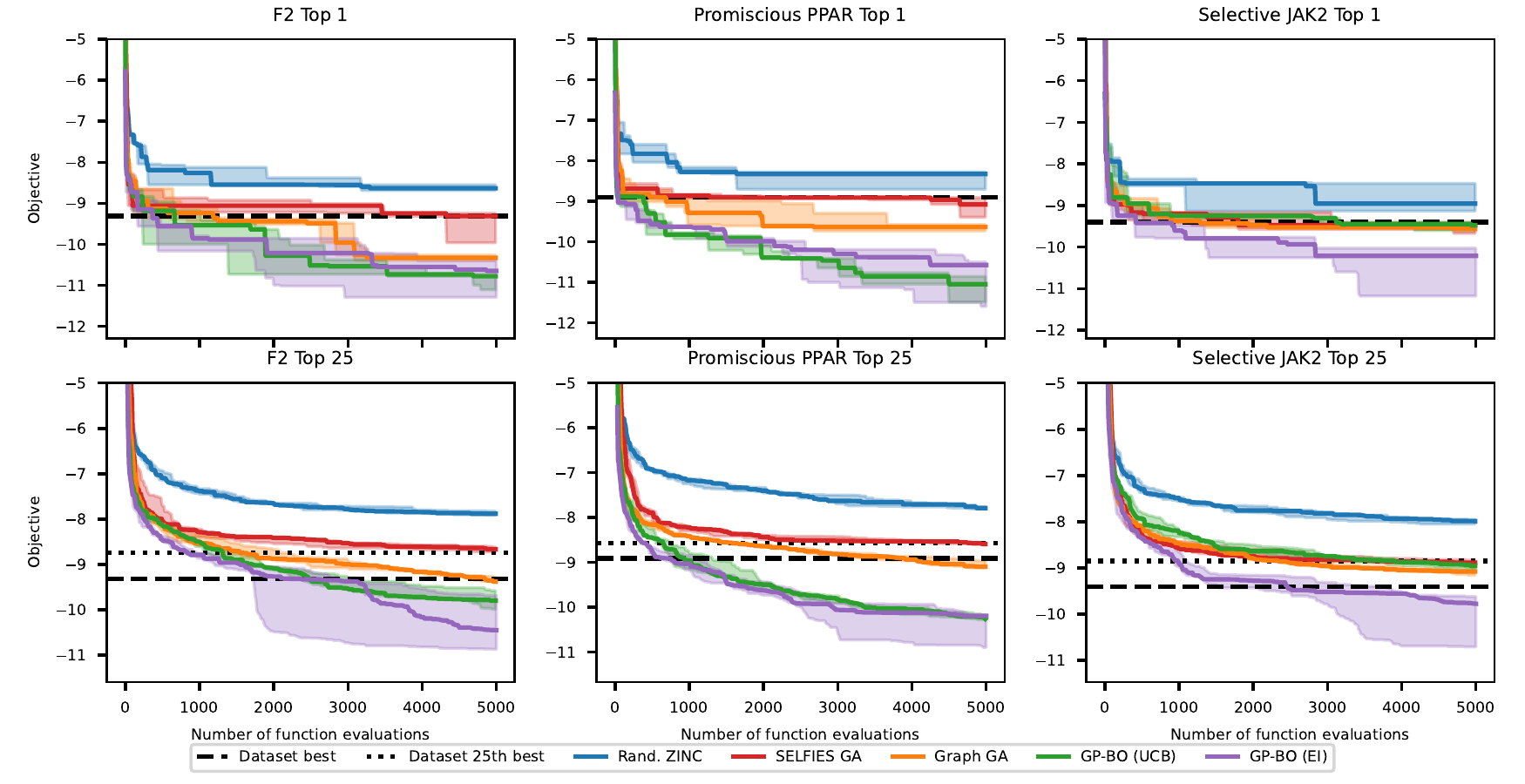}
	\caption{
		Results for baseline algorithms on three different \textit{de novo} molecular design tasks.
		The objective value of the first and 25th best molecule found so far is shown as a function of the number of objective function calls.
		The solid lines indicate the median and the shaded area the minimum and maximum over three runs.
		The black dashed line indicates the best (and 25th best) value in the \dockstring{} dataset.
	}
	\label{fig:molopt-top1-top25-trajectories}
\end{figure}

The optimization trajectories of the penalized tasks (Figure~\ref{fig:molopt-top1-top25-trajectories}, top) were generally flatter than that of the naive unpenalized one, suggesting that they are more difficult. 
In F2, three of the methods beat the best molecules in the dataset by a large margin, compared with two methods for Promiscuous PPAR and just one method for Selective JAK2, suggesting that the task difficulty increases in that order. 
In general the GP-BO algorithms tend to significantly outperform the GAs, although GP-BO with UCB acquisition is comparable to the GAs for the selective JAK2 task. 
Random sampling of ZINC molecules yielded the worst performance, which is expected since this strategy does not learn from past molecules unlike other optimization methods. 
The objective value of the 25th best molecule so far (Figure~\ref{fig:molopt-top1-top25-trajectories}, bottom) showed a similar relative performance of optimization algorithms as in the single best molecule, except that differences between algorithms were more pronounced. 
In addition, only a single method, GP-BO with EI acquisition, was able to find 25th best molecules better than the training set in all tasks. This suggests that finding multiple high-performing molecules is more challenging than finding a single {high-performing} molecule, as expected.

Molecules generated in F2 and Promiscious PPAR featured conjugated ring structures which are relatively unusual in successful drugs (Figure~\ref{fig:all-tasks-topmols}, second and third row).
%\mg{Ask Andreas why conjugated ring structures are unusual if their \gls{qed} is high. That should mean that they are usual}
Selective JAK2 yielded smaller molecules, with interesting structures, druglike appearance, and higher \gls{qed} values, although all the top molecules shared a similar backbone (Figure~\ref{fig:all-tasks-topmols}, fourth row).
We hypothesize that adversarial objectives based on correlated docking scores may be an effective way to avoid docking biases compared to simple penalties based on \gls{qed},
since exploiting the bias of docking scores for high molecular size and lipophilicity may benefit one component of the objective while hurting another.
Future work is needed to further study and verify this effect.

\begin{figure}[ht]
    %\vspace{0.5cm}
	\centering
	\includegraphics[width=\textwidth]{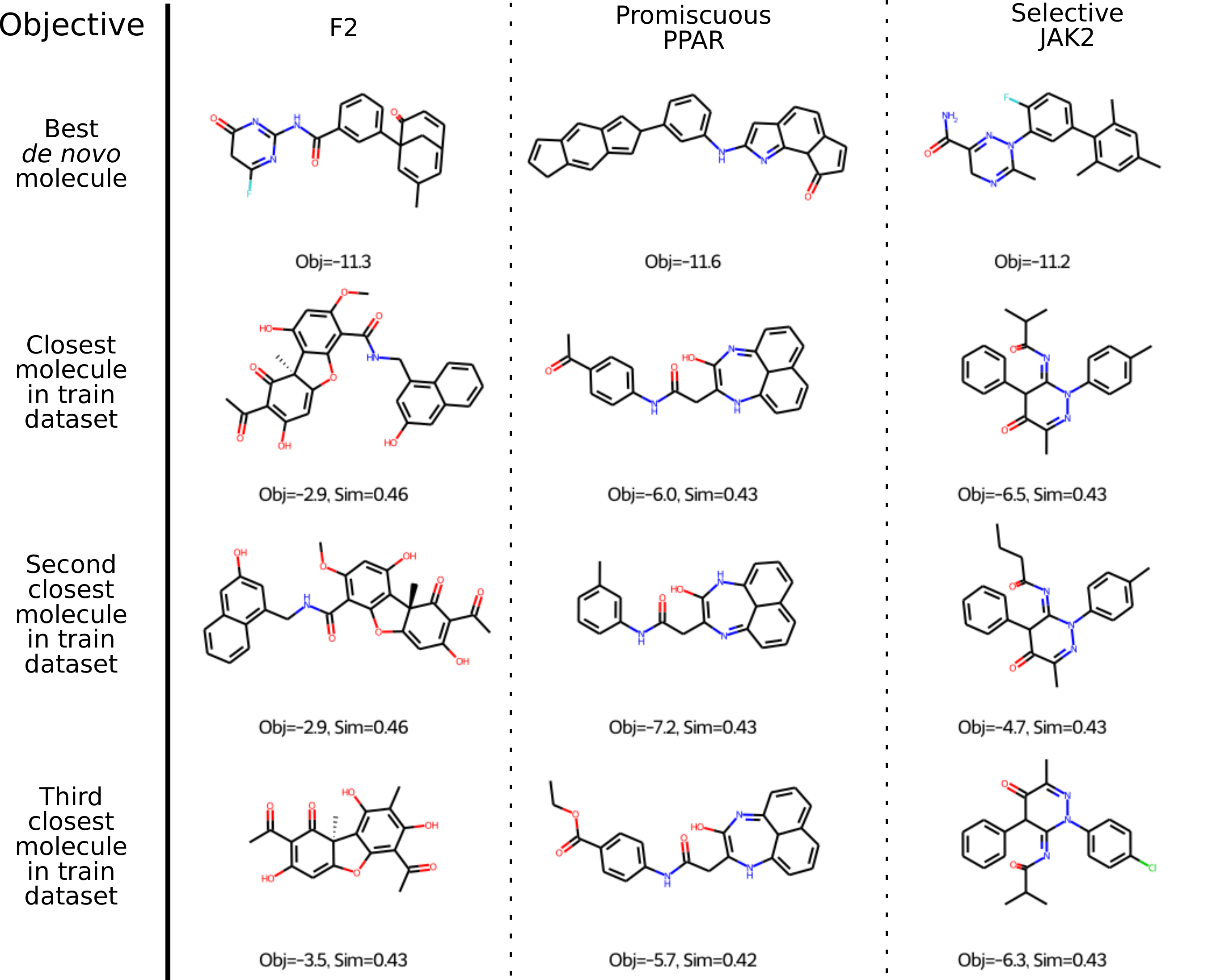}
	%\vspace{0.5cm}
	\caption{
	    Most similar molecules in training set for the three objectives F2, Promiscuous PPAR, and Selective JAK2.
    }
	\label{fig:molopt-train-sim}
\end{figure}

In general, the best molecules in the three tasks are unique and distinct from the training set (Figure~\ref{fig:molopt-train-sim}).
For F2 and Promiscuous PPAR, none of the top molecules has a generic Murcko scaffold in the training set; 
for Selective JAK2, all of the top 12 molecules share a generic Murcko scaffold with a training set molecule but the most similar molecule is still reasonably different.

\vspace{1cm}

To compare the difficulty of our \textit{de novo} design tasks with other popular benchmark functions, we assessed the performance of two baseline models when optimizing logP and \gls{qed}.
Our results suggest that neither logP nor \gls{qed} are appropriate objectives for model evaluation. 
LogP was remarkably easy to optimize for all methods, in line with previous regression results indicating that this property is not challenging enough (cf.\ Table~\ref{tab:regression-shortlist}).
Furthermore, it promoted molecules that were highly unrealistic and not druglike (Figure~\ref{fig:logP-topmols} of the \gls{si}).
On the other hand, \gls{qed} seemed to be maximized by molecules already in the dataset and it could not be improved further than  $0.948$.
Since \gls{qed} is itself a scalarized multi-objective function of several physicochemical properties, this suggests that many existing molecules in chemical depositories are already in the \gls{qed} Pareto frontier.
%\gncs{I don't understand what this Pareto frontier is doing here.}
Therefore, \gls{qed} may be more useful as a soft constraint for druglikeness (as employed in this work) than as a benchmark objective.

%%%%%%%%%%%%%%%%%%%%%%%%%%%%%%%%%%%%%%%%%%%%%%%%%%%%%%%%%%%%
\section{Conclusions and Outlook}
\label{sec:conclusion}
%%%%%%%%%%%%%%%%%%%%%%%%%%%%%%%%%%%%%%%%%%%%%%%%%%%%%%%%%%%%

With the release of \dockstring{} we hope to make docking-based benchmarking as accessible as possible, and thus, 
enable the scientific community to benchmark algorithms against challenging and relevant tasks in drug discovery.
The simple and robust Python package enables automatic computation of docking scores and poses, facilitating the acquisition of new labels and the design of sophisticated workflows of virtual screening or molecular optimization---even by researchers with little domain expertise.
The dataset of unprecedented size and diversity allows users to train models without having to spend significant computational resources. 
Furthermore, it provides curated and standardized training and test sets for each benchmark so that models are compared fairly.
This consideration is particularly important to ML for chemistry, given that different dataset splits can lead to largely disparate results due to the biased and undersampled nature of chemical space.
Our training and test sets were constructed with cluster splitting, to minimize the chances of overfitting and data leakage.
Finally, the set of benchmark tasks is carefully designed so that they are relevant to both the ML and the drug discovery communities,
covering a variety of ML settings and biological problems.

% Maintenance plan
The possibilities for tasks based on docking are by no means exhausted in this paper,
and we plan to continue improving the package, dataset, and benchmarks
(see Section~\ref{sec:maintenance} of the \gls{si} for our maintenance plan).
The following areas are of particular interest.
% Improve de novo tasks
First, there is room to adapt and improve the \textit{de novo} design tasks, in particular the objective functions, to encourage the generation of molecules with better pharmacokinetic properties and more feasible synthetic pathways.
% Transfer learning
Secondly, the range of protein targets included in \dockstring{} makes it well suited to multi-objective tasks such as transfer learning, self-supervised learning, and few-shot learning. These are left for future work.
% Multi-fidelity
Thirdly, docking scores are considered a relatively limited predictor of bioactivity, because, among other reasons,
they use a static binding site and force fields which are poorly calibrated for certain metal ions, for instance.
Therefore, drug discovery projects tend to employ more expensive computational techniques and experimental assays in later stages of the drug discovery pipeline.
Developing transfer learning and multi-fidelity optimization tasks for different predictors of activity on the same \dockstring{} target would be a relevant avenue of future research.

%%%%%%%%%%%%%%%%%%%%%%%%%%%%%%%%%%%%%%%%%%%%%%%%%%%%%%%%%%%%
\section*{Data and Software Availability}
\label{sec:availability}
%%%%%%%%%%%%%%%%%%%%%%%%%%%%%%%%%%%%%%%%%%%%%%%%%%%%%%%%%%%%

The \dockstring{} molecular docking package is available at \githuburl{}. The \dockstring{} dataset, as well as code for the baselines, are available at \figshareurl{}. All components are released under the Apache 2.0 license.

%%%%%%%%%%%%%%%%%%%%%%%%%%%%%%%%%%%%%%%%%%%%%%%%%%%%%%%%%%%%
% Acknowledgements
%%%%%%%%%%%%%%%%%%%%%%%%%%%%%%%%%%%%%%%%%%%%%%%%%%%%%%%%%%%%

\begin{acknowledgement}
	MGO  acknowledges support from a Wellcome Trust Doctoral Studentship.
	GNCS and JMHL acknowledge support from a Turing AI Fellowship under grant EP/V023756/1.
	AJT acknowledges funding via a C T Taylor Cambridge International Scholarship.
	SB acknowledges support from MRC grant MR/P01710X/1.
	This work has been performed using resources operated by the University of Cambridge Research Computing Service,
	which is funded by the EPSRC (capital grant EP/P020259/1) and DiRAC funding from the STFC (\url{http://www.dirac.ac.uk/}).
\end{acknowledgement}

%%%%%%%%%%%%%%%%%%%%%%%%%%%%%%%%%%%%%%%%%%%%%%%%%%%%%%%%%%%%
% References
%%%%%%%%%%%%%%%%%%%%%%%%%%%%%%%%%%%%%%%%%%%%%%%%%%%%%%%%%%%%

\bibliography{references}

%%%%%%%%%%%%%%%%%%%%%%%%%%%%%%%%%%%%%%%%%%%%%%%%%%%%%%%%%%%%
\clearpage
\newpage
\appendix
% \counterwithin{figure}{section}
\setcounter{figure}{0}
\setcounter{table}{0}
% \counterwithin{table}{section}
%%%%%%%%%%%%%%%%%%%%%%%%%%%%%%%%%%%%%%%%%%%%%%%%%%%%%%%%%%%%

\begin{center}
	\begin{LARGE}
		\bf Supporting Information
	\end{LARGE}
\end{center}

%\gncs{We need the manuscript's title here.}
\vspace{1cm}

%%%%%%%%%%%%%%%%%%%%%%%%%%%%%%%%%%%%%%%%%%%%%%%%%%%%%%%%%%%%
\section{Popular Molecular Benchmarks and Their Relevance to Drug Discovery}
\label{app:bad_benchmarks}
%%%%%%%%%%%%%%%%%%%%%%%%%%%%%%%%%%%%%%%%%%%%%%%%%%%%%%%%%%%%

%\begin{noindent}
\begin{table}[h!]
	\centering
	\caption{
		Popular molecular benchmarks and their relevance to drug discovery.
	}
	\label{tab:current_benchmarks_are_bad}
	\begin{tabular}{@{}p{2.0cm}p{5.5cm}p{6.5cm}@{}}
		\toprule
		& Description      & Relevance for drug discovery   \\
		\midrule
		logP             
		&  
		Ratio of the concentrations of a compound in a mixture of an organic solvent and water. 
		&  
		Some heuristic rules consider the logP (e.g., less than five in Lipinski's rule of five \cite{Lipinski2001Experimental}) since molecules with a high logP often suffer from unspecific binding and safety liabilities.
		\\
		\gls{qed}
		&
		The \textbf{Q}uantitative \textbf{E}stimate of \textbf{D}ruglikeness measures similarity to marketed drugs based on simple physicochemical properties.
		&
		\gls{qed} has limited predictive power to discriminate approved drugs from decoys \cite{yusof2013considering}. 
		Further, it is not selective against a particular disease or target protein.
		\\
		SAS
		&  Based on the similarity to synthesizable compounds,
		the \textbf{S}ynthetic \textbf{A}ccessibility \textbf{S}core estimates how difficult it is to synthesize a molecule.
		&
		Synthesizability is a pre-requisite for any molecule to be assayed \textit{in vitro} or \textit{in vivo}, 
		but it does not inform about efficacy or safety.
		\\
		Molecular mass
		&
		Often referred to as the molecular weight.
		&   
		While some heuristic rules for druglikeness consider molecular weight (e.g., less than 500 Da in Lipinski's rule of five), 
		it offers little information about efficacy and safety.
		\\
		Docking scores
		&
		Prediction of binding free energy between a molecule (ligand) and a protein (target).
		&
		Popular method in virtual screening with the goal to enrich a subset with bioactive compounds from an extensive molecular library.
		\\
		\bottomrule
	\end{tabular}
\end{table}
%\end{noindent}

\clearpage
\newpage

%%%%%%%%%%%%%%%%%%%%%%%%%%%%%%%%%%%%%%%%%%%%%%%%%%%%%%%%%%%%
\section{\dockstring{} Target Proteins}
\label{app:target_proteins}
%%%%%%%%%%%%%%%%%%%%%%%%%%%%%%%%%%%%%%%%%%%%%%%%%%%%%%%%%%%%

%\begin{noindent}
\begin{table}[h!]
	\centering
	\caption{
        Examples of high-quality targets in the \dockstring{} dataset that are relevant for drug discovery.
        Drugs in this table are all small molecules \cite{wishart2006drugBank}.
	}
	\label{tab:drugs-approved}
	\begin{tabular}{@{} P{4.5cm} P{2cm} P{4.5cm} P{4.5cm}@{}}
		\toprule
		Target            & Group & Biological significance & Drug                                                            \\
		\midrule
		Janus kinase 2 (JAK2)           & Kinase     & Cell signaling via the JAK-STAT pathway. Misregulated or mutated in a range of cancers. & Ruxolitinib. Selective JAK inhibitor used to treat myelofibrosis, a rare type of bone marrow blood cancer.  \\
		Tyrosine-protein kinase (KIT)           & Kinase     & Cell-surface receptor and signal transducer for cytokines. Misregulated or mutated in a range of cancers. & Axitinib. Used to treat renal cell carcinoma.  \\
		Hepatocyte growth factor receptor (MET)           & Kinase     & Cell-surface receptor and signal transducer for the hepatocyte growth factor. Initiator of the MET signaling pathway. Misregulated or mutated in a range of cancers. & Crizotinib. Used for the treatment of non-small cell lung cancer (NSCLC).  \\
		Thrombin (F2)           & Protease     &   Catalizes the cleavage of soluble fibrinogen into insoluble fibrin to promote blood coagulation during blood clotting.  & Bivalirudin. Anticoagulant used to prevent thrombosis in patients under heparin treatment. \\
                Peroxisome proliferator-activated receptor alpha (PPARA)  &  Nuclear receptor   &    Regulator of liver metabolism. Activates uptake and utilization of fatty acids.   &   Clofibrate. Used to control high cholesterol and triglyceride levels in the blood. \\
                Cyclic guanosine monophosphate specific phosphodiesterase type 5 (PDE5A)  &  Enzyme   &    Degrades the messenger cGMP, promoting vasodilation and increased blood flow.   &   Sildenafil (viagra). Used to treat erectile dysfunction. \\
		\bottomrule
	\end{tabular}
\end{table}
%\end{noindent}

\clearpage
\newpage

%%%%%%%%%%%%%%%%%%%%%%%%%%%%%%%%%%%%%%%%%%%%%%%%%%%%%%%%%%%%
\section{Benchmark Details}
\label{apdx:benchmark-details}
%%%%%%%%%%%%%%%%%%%%%%%%%%%%%%%%%%%%%%%%%%%%%%%%%%%%%%%%%%%%

Code for all baselines is provided at \figshareurl{}.

%%%%%%%%%%%%%%%%%%%%%%%%%%%%%%%%%%%%%%%%%%%%%%%%%%%%%%%%%%%%
\subsection{Regression}
\label{apdx:regression-details}
%%%%%%%%%%%%%%%%%%%%%%%%%%%%%%%%%%%%%%%%%%%%%%%%%%%%%%%%%%%%

\subsubsection{Additional Details}

\paragraph{Clipping positive scores.}
Docking scores are clipped to a maximum value of $+5$ before fitting the
regression model
because there were a small number of huge positive scores (e.g.,\@$+100$),
which we worried would have a large negative impact on the training of some models.
Positive docking scores represent poor binding and therefore predicting
the exact value of a positive docking score is uninteresting.
For most targets, there were no positive scores and therefore this clipping had no effect.

\paragraph{Train/test split.}
Specifically, the train/test sets were produced by sorting the clusters
by size, then adding all the molecules in the largest cluster to the training set.
This process was repeated using the remaining clusters until the 85\% of the dataset had been added to the training set.
The remaining data points were used as the test set,
and consist mostly of small, isolated clusters.

%%%%%%%%%%%%%%%%%%%%%%%%%%%%%%%%%%%%%%%%%%%%%%%%%%%%%%%%%%%%
\subsubsection{Results}
%%%%%%%%%%%%%%%%%%%%%%%%%%%%%%%%%%%%%%%%%%%%%%%%%%%%%%%%%%%%

\paragraph{$R^2$ score.}
There are several similar definitions of $R^2$ score.
The one used here is the implementation from \texttt{scikit-learn} in
\url{https://scikit-learn.org/stable/modules/generated/sklearn.metrics.r2_score.html}.
With this definition, perfect prediction gets a score of 1.0, while predicting the dataset mean will have a score of 0.0.

\begin{table}[ht]
	\centering
	\caption{
		Coefficient of determination ($R^2$, higher is better) for regression baseline methods.
		Results shown here are the mean of three runs;
		the full table with standard deviations can be found in the Appendix (Table~\ref{tab:regression-full}).
		All standard deviations were small.
		The vertical line separates classical fingerprint-based methods from deep learning methods.
		The best score in each row is in \textbf{bold}.
	}
	\label{tab:regression}

	\resizebox{\columnwidth}{!}{%
		\begin{tabular}{lccccccc}
			\toprule
			Target                & Ridge & Lasso & XGBoost & GP (exact) & GP (sparse) & MPNN  & Attentive FP   \\
			\midrule
			logP                  & 0.640 & 0.640 & 0.734   & 0.707      & 0.716       & 0.953 & \textbf{1.000} \\
			QED                   & 0.519 & 0.483 & 0.660   & 0.640      & 0.598       & 0.901 & \textbf{0.981} \\
			\midrule
			ADAM17                & 0.597 & 0.591 & 0.661   & 0.638      & 0.685       & 0.748 & \textbf{0.822} \\
			ESR1                  & 0.499 & 0.478 & 0.567   & 0.527      & 0.584       & 0.609 & \textbf{0.729} \\
			ESR2                  & 0.421 & 0.416 & 0.497   & 0.441      & 0.508       & 0.506 & \textbf{0.627} \\
			F10                   & 0.655 & 0.652 & 0.685   & 0.687      & 0.727       & 0.743 & \textbf{0.856} \\
			F2                    & 0.672 & 0.663 & 0.688   & 0.705      & 0.744       & 0.798 & \textbf{0.880} \\
			HSD11B1               & 0.425 & 0.424 & 0.577   & 0.542      & 0.612       & 0.620 & \textbf{0.793} \\
			IGF1R                 & 0.583 & 0.574 & 0.632   & 0.615      & 0.666       & 0.685 & \textbf{0.799} \\
			JAK2                  & 0.617 & 0.617 & 0.686   & 0.667      & 0.712       & 0.759 & \textbf{0.853} \\
			KIT                   & 0.604 & 0.594 & 0.674   & 0.637      & 0.684       & 0.755 & \textbf{0.806} \\
			LCK                   & 0.670 & 0.666 & 0.706   & 0.708      & 0.746       & 0.789 & \textbf{0.890} \\
			MAPK14                & 0.592 & 0.587 & 0.655   & 0.640      & 0.689       & 0.731 & \textbf{0.814} \\
			MAPKAPK2              & 0.662 & 0.660 & 0.704   & 0.699      & 0.736       & 0.788 & \textbf{0.855} \\
			MET                   & 0.587 & 0.584 & 0.663   & 0.633      & 0.683       & 0.766 & \textbf{0.804} \\
			NR3C1                 & 0.258 & 0.257 & 0.495   & 0.439      & 0.520       & 0.425 & \textbf{0.753} \\
			PARP1                 & 0.706 & 0.700 & 0.723   & 0.743      & 0.772       & 0.815 & \textbf{0.910} \\
			PDE5A                 & 0.601 & 0.600 & 0.661   & 0.663      & 0.702       & 0.769 & \textbf{0.851} \\
			PGR                   & 0.242 & 0.245 & 0.345   & 0.291      & 0.387       & 0.324 & \textbf{0.678} \\
			PPARA                 & 0.577 & 0.575 & 0.645   & 0.626      & 0.675       & 0.736 & \textbf{0.823} \\
			PPARD                 & 0.630 & 0.627 & 0.686   & 0.667      & 0.714       & 0.782 & \textbf{0.851} \\
			PPARG                 & 0.641 & 0.634 & 0.677   & 0.662      & 0.707       & 0.770 & \textbf{0.818} \\
			PTGS2                 & 0.322 & 0.310 & 0.398   & 0.349      & 0.419       & 0.427 & \textbf{0.588} \\
			PTK2                  & 0.611 & 0.603 & 0.675   & 0.657      & 0.700       & 0.751 & \textbf{0.839} \\
			PTPN1                 & 0.600 & 0.596 & 0.660   & 0.630      & 0.677       & 0.706 & \textbf{0.790} \\
			SRC                   & 0.663 & 0.660 & 0.689   & 0.692      & 0.735       & 0.802 & \textbf{0.875} \\
			\midrule
			\textbf{Average Rank} & 6.042 & 6.958 & 4.167   & 4.708      & 2.875       & 2.250 & \textbf{1.000} \\
			\bottomrule
		\end{tabular}
	}
\end{table}

\newpage

\begin{landscape}
	\begin{table}
		\centering
		\caption{Full regression results (complete version of Table~\ref{tab:regression} with standard deviations.}
		\label{tab:regression-full}

		%\begin{noindent}
	\resizebox{0.85\columnwidth}{!}{%
    \begin{tabular}{lcccccccccccccc}
    \toprule
    {} & \multicolumn{2}{l}{Ridge} & \multicolumn{2}{l}{Lasso} & \multicolumn{2}{l}{XGBoost} & \multicolumn{2}{l}{GP (exact)} & \multicolumn{2}{l}{GP (sparse)} & \multicolumn{2}{l}{MPNN} & \multicolumn{2}{l}{Attentive FP} \\
    Target   &   mean &    std &   mean &    std &    mean &    std &       mean &    std &        mean &    std &   mean &    std &            mean &    std \\
    \midrule
    logP     &  0.640 &  0.000 &  0.640 &  0.000 &   0.734 &  0.000 &      0.707 &  0.003 &       0.716 &  0.007 &  0.953 &  0.007 &  \textbf{1.000} &  0.000 \\
    QED      &  0.519 &  0.000 &  0.483 &  0.008 &   0.660 &  0.000 &      0.640 &  0.003 &       0.598 &  0.053 &  0.901 &  0.006 &  \textbf{0.981} &  0.001 \\
    \hline
    ADAM17   &  0.597 &  0.000 &  0.591 &  0.009 &   0.661 &  0.000 &      0.638 &  0.003 &       0.685 &  0.001 &  0.748 &  0.031 &  \textbf{0.822} &  0.006 \\
    ESR1     &  0.499 &  0.000 &  0.478 &  0.016 &   0.567 &  0.000 &      0.527 &  0.002 &       0.584 &  0.001 &  0.609 &  0.016 &  \textbf{0.729} &  0.002 \\
    ESR2     &  0.421 &  0.001 &  0.416 &  0.008 &   0.497 &  0.000 &      0.441 &  0.002 &       0.508 &  0.000 &  0.506 &  0.001 &  \textbf{0.627} &  0.010 \\
    F10      &  0.655 &  0.000 &  0.652 &  0.002 &   0.685 &  0.000 &      0.687 &  0.001 &       0.727 &  0.000 &  0.743 &  0.028 &  \textbf{0.856} &  0.001 \\
    F2       &  0.672 &  0.000 &  0.663 &  0.009 &   0.688 &  0.000 &      0.705 &  0.002 &       0.744 &  0.000 &  0.798 &  0.005 &  \textbf{0.880} &  0.001 \\
    HSD11B1  &  0.425 &  0.000 &  0.424 &  0.001 &   0.577 &  0.000 &      0.542 &  0.003 &       0.612 &  0.002 &  0.620 &  0.023 &  \textbf{0.793} &  0.006 \\
    IGF1R    &  0.583 &  0.000 &  0.574 &  0.010 &   0.632 &  0.000 &      0.615 &  0.002 &       0.666 &  0.001 &  0.685 &  0.037 &  \textbf{0.799} &  0.003 \\
    JAK2     &  0.617 &  0.000 &  0.617 &  0.000 &   0.686 &  0.000 &      0.667 &  0.001 &       0.712 &  0.002 &  0.759 &  0.006 &  \textbf{0.853} &  0.003 \\
    KIT      &  0.604 &  0.000 &  0.594 &  0.011 &   0.674 &  0.000 &      0.637 &  0.002 &       0.684 &  0.001 &  0.755 &  0.005 &  \textbf{0.806} &  0.008 \\
    LCK      &  0.670 &  0.000 &  0.666 &  0.002 &   0.706 &  0.000 &      0.708 &  0.001 &       0.746 &  0.000 &  0.789 &  0.026 &  \textbf{0.890} &  0.000 \\
    MAPK14   &  0.592 &  0.000 &  0.587 &  0.006 &   0.655 &  0.000 &      0.640 &  0.001 &       0.689 &  0.001 &  0.731 &  0.010 &  \textbf{0.814} &  0.004 \\
    MAPKAPK2 &  0.662 &  0.000 &  0.660 &  0.001 &   0.704 &  0.000 &      0.699 &  0.002 &       0.736 &  0.001 &  0.788 &  0.031 &  \textbf{0.855} &  0.010 \\
    MET      &  0.587 &  0.000 &  0.584 &  0.004 &   0.663 &  0.000 &      0.633 &  0.004 &       0.683 &  0.001 &  0.766 &  0.011 &  \textbf{0.804} &  0.009 \\
    NR3C1    &  0.258 &  0.002 &  0.257 &  0.001 &   0.495 &  0.000 &      0.439 &  0.003 &       0.520 &  0.002 &  0.425 &  0.030 &  \textbf{0.753} &  0.005 \\
    PARP1    &  0.706 &  0.000 &  0.700 &  0.004 &   0.723 &  0.000 &      0.743 &  0.002 &       0.772 &  0.002 &  0.815 &  0.010 &  \textbf{0.910} &  0.002 \\
    PDE5A    &  0.601 &  0.000 &  0.600 &  0.000 &   0.661 &  0.000 &      0.663 &  0.002 &       0.702 &  0.004 &  0.769 &  0.013 &  \textbf{0.851} &  0.003 \\
    PGR      &  0.242 &  0.002 &  0.245 &  0.001 &   0.345 &  0.000 &      0.291 &  0.007 &       0.387 &  0.000 &  0.324 &  0.096 &  \textbf{0.678} &  0.008 \\
    PPARA    &  0.577 &  0.000 &  0.575 &  0.002 &   0.645 &  0.000 &      0.626 &  0.002 &       0.675 &  0.001 &  0.736 &  0.017 &  \textbf{0.823} &  0.008 \\
    PPARD    &  0.630 &  0.000 &  0.627 &  0.001 &   0.686 &  0.000 &      0.667 &  0.003 &       0.714 &  0.001 &  0.782 &  0.010 &  \textbf{0.851} &  0.003 \\
    PPARG    &  0.641 &  0.000 &  0.634 &  0.007 &   0.677 &  0.000 &      0.662 &  0.002 &       0.707 &  0.001 &  0.770 &  0.010 &  \textbf{0.818} &  0.003 \\
    PTGS2    &  0.322 &  0.001 &  0.310 &  0.017 &   0.398 &  0.000 &      0.349 &  0.004 &       0.419 &  0.003 &  0.427 &  0.004 &  \textbf{0.588} &  0.020 \\
    PTK2     &  0.611 &  0.000 &  0.603 &  0.003 &   0.675 &  0.000 &      0.657 &  0.003 &       0.700 &  0.002 &  0.751 &  0.013 &  \textbf{0.839} &  0.002 \\
    PTPN1    &  0.600 &  0.000 &  0.596 &  0.005 &   0.660 &  0.000 &      0.630 &  0.001 &       0.677 &  0.001 &  0.706 &  0.014 &  \textbf{0.790} &  0.002 \\
    SRC      &  0.663 &  0.000 &  0.660 &  0.001 &   0.689 &  0.000 &      0.692 &  0.002 &       0.735 &  0.001 &  0.802 &  0.013 &  \textbf{0.875} &  0.002 \\
    \bottomrule
    \end{tabular}
    }
    %\end{noindent}

	\end{table}
\end{landscape}

\newpage

%%%%%%%%%%%%%%%%%%%%%%%%%%%%%%%%%%%%%%%%%%%%%%%%%%%%%%%%%%%%
\subsection{Virtual Screening}
%%%%%%%%%%%%%%%%%%%%%%%%%%%%%%%%%%%%%%%%%%%%%%%%%%%%%%%%%%%%

%%%%%%%%%%%%%%%%%%%%%%%%%%%%%%%%%%%%%%%%%%%%%%%%%%%%%%%%%%%%
\subsubsection{Additional task details.}
%%%%%%%%%%%%%%%%%%%%%%%%%%%%%%%%%%%%%%%%%%%%%%%%%%%%%%%%%%%%

\paragraph{Training.}
All training details are identical to those presented in Section~\ref{apdx:regression-details}.

\paragraph{ZINC dataset.}
Because the ZINC dataset grows over time,
a copy of the dataset was downloaded from \url{https://zinc20.docking.org/}
in July 2021 to be used as the standard dataset for this task.
It contained 997597004 SMILES strings.
There were 56606 items in common between ZINC and the training set,
representing 22\% of the training set.

\paragraph{Enrichment factor calculation.}
Because the true docking scores of all compounds in the ZINC dataset are too expensive to calculate, 
the exact cutoff for the top 0.1\% of the dataset is unknown.
To estimate it, we selected a random subset of ZINC20 of size 100K for each target
and calculated the scores with \dockstring{}.
The top 0.1\% of this subset was used as an estimate of the true 0.1\% cutoff.

\paragraph{Threshold for enrichment factor.}
The threshold of 0.1\% was chosen for two reasons.
First, it has been given as the approximate hit rate of high-throughput screening \cite{bender2008which}.
Second, if the threshold were higher (say top 1\%), the task would not be as challenging,
and the differences in methods might not be as apparent.
Third, the 0.1\% threshold is estimated from a sample size of 100,000,
making it the docking score of the 100\textsuperscript{th} best molecule in the sample.
If the percentile were much lower, the estimate of the cutoff value might be unreliable.

%%%%%%%%%%%%%%%%%%%%%%%%%%%%%%%%%%%%%%%%%%%%%%%%%%%%%%%%%%%%
\subsubsection{Results.}
%%%%%%%%%%%%%%%%%%%%%%%%%%%%%%%%%%%%%%%%%%%%%%%%%%%%%%%%%%%%

\paragraph{How good are the docking scores?}
For KIT, ridge finds one, and Attentive FP finds 35 molecules with docking scores lower than the lowest in the training set.
For PARP1, ridge finds one, and Attentive FP finds ten molecules with docking scores lower than the lowest in the training set.
For PGR, ridge and Attentive FP find zero molecules with docking scores lower than the lowest in the training set.

\paragraph{How do the results depend on the threshold for active molecules?}
If the threshold is increased (e.g.,\@ top 1\%), the results are qualitatively the same, although the quantitative differences are less pronounced.
Lowering the threshold has the opposite effect.

%%%%%%%%%%%%%%%%%%%%%%%%%%%%%%%%%%%%%%%%%%%%%%%%%%%%%%%%%%%%
\subsection{\textit{De Novo} Molecular Design}
\label{apdx:molopt-details}
%%%%%%%%%%%%%%%%%%%%%%%%%%%%%%%%%%%%%%%%%%%%%%%%%%%%%%%%%%%%

\subsubsection{Baseline methods.}

\paragraph{Rationale for choice of baselines.}
We selected the methods to represent two broad classes of algorithms used in previous work.
% GAs
Genetic algorithms are commonly used for molecular design:
the graph genetic algorithm \cite{jensen2019a} was chosen due to its strong performance in the GuacaMol baselines \cite{brown2019guacamol},
while the SELFIES genetic algorithm was chosen due to its simplicity.
We believe these algorithms are representative of a broader class of model-free, exploratory algorithms.
% BO
Bayesian optimization with a \gls{gp} was chosen because Bayesian optimization is widely regarded as a high-performing
optimization technique when the number of function evaluations is limited \cite{shahriari2015taking}.
\Glspl{gp} are the most common model used in Bayesian optimization.
The acquisition functions chosen are the most commonly employed acquisition functions
as far as we are aware.
We believe that GP-BO represents a broader class of model-based algorithms that can be used
for both exploration and exploitation.
% Random
Finally, random ZINC is an important trivial baseline which acts as a lower bound
for acceptable performance for an algorithm.

\paragraph{Maximization or minimization?}
Although most of the objectives in Section~\ref{ssec:molopt} are minimization objectives,
our code was designed for maximization.
Therefore all minimization objectives are multiplied by $-1$ and maximized.
Technical descriptions in the remainder of this section therefore correspond to maximization.

\paragraph{Reinforcement learning.}
We omitted any baseline methods based on reinforcement learning.
This is because, to our knowledge, previously reported policy reinforcement learning
methods required many more than 5000 objective function evaluations to achieve reasonable performance (e.g.,\@ Ref.~\citenum{you2018graph}).
Reinforcement learning will be explored in future version of this manuscript.

\clearpage 

%%%%%%%%%%%%%%%%%%%%%%%%%%%%%%%%%%%%%%%%%%%%%%%%%%%%%%%%%%%%
\subsubsection{Results}
%%%%%%%%%%%%%%%%%%%%%%%%%%%%%%%%%%%%%%%%%%%%%%%%%%%%%%%%%%%%

\begin{figure}[h!]
	\center
	\centering
	\begin{tabular}{lcccccc}
\toprule
{} &  Objective &   logP &  Molecular Weight &  HBA &  HBD &   QED \\
Rank &            &        &                   &      &      &       \\
\midrule
1    &     43.268 & 43.268 &          3069.380 &   36 &    0 & 0.014 \\
2    &     42.629 & 42.629 &          3190.526 &   33 &    0 & 0.014 \\
3    &     42.486 & 42.486 &          3157.732 &   33 &    0 & 0.014 \\
4    &     42.439 & 42.439 &          3068.325 &   36 &    0 & 0.014 \\
5    &     41.997 & 41.997 &          3188.366 &   37 &    0 & 0.014 \\
6    &     41.960 & 41.960 &          3193.075 &   33 &    0 & 0.014 \\
%7    &     41.611 & 41.611 &          2692.912 &   22 &    0 & 0.011 \\
%8    &     41.524 & 41.524 &          2929.261 &   33 &    0 & 0.014 \\
%9    &     41.348 & 41.348 &          2891.033 &   22 &    0 & 0.014 \\
%10   &     41.011 & 41.011 &          2857.325 &   35 &    0 & 0.014 \\
%11   &     40.856 & 40.856 &          3200.145 &   32 &    0 & 0.014 \\
%12   &     40.808 & 40.808 &          2958.244 &   18 &    0 & 0.014 \\
\bottomrule
\end{tabular}

	\includegraphics[width=1.1\textwidth]{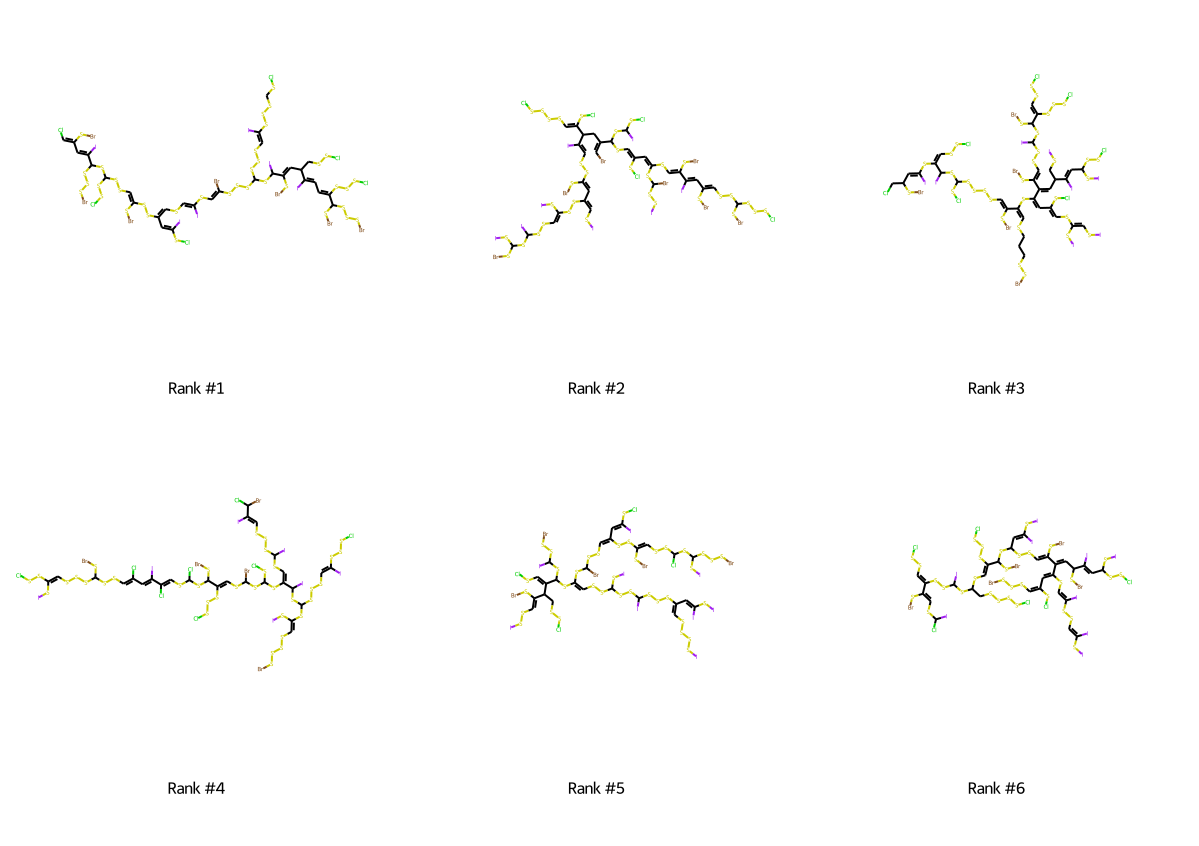}
	\caption{
		Best molecules for logP optimization.
	}
	\label{fig:logP-topmols}
\end{figure}

\clearpage
\newpage

%%%%%%%%%%%%%%%%%%%%%%%%%%%%%%%%%%%%%%%%%%%%%%%%%%%%%%%%%%%%
\section{Computational Details}
\label{sec:comp_details}
%%%%%%%%%%%%%%%%%%%%%%%%%%%%%%%%%%%%%%%%%%%%%%%%%%%%%%%%%%%%

%%%%%%%%%%%%%%%%%%%%%%%%%%%%%%%%%%%%%%%%%%%%%%%%%%%%%%%%%%%%
\subsection{Docking Run Time}
%%%%%%%%%%%%%%%%%%%%%%%%%%%%%%%%%%%%%%%%%%%%%%%%%%%%%%%%%%%%

Average \dockstring{} docking run times are shown in Table~\ref{tab:docking-times}.

\begin{table}
	\caption{
		Mean (and standard deviation) of docking times for all targets with \dockstring{},
		averaged over 50 ligands from the \dockstring{} dataset.
		Results are shown for different numbers of CPUs.
		The same set of 50 ligands was used for every target.
		Continued in Table~\ref{tab:docking-times2}.
	}
	\label{tab:docking-times}
	\begin{tabular}{lcccccc}
		\toprule
		Target   & 1 CPU         & 2 CPUs      & 4 CPUs      & 8 CPUs      & 16 CPUs    & 32 CPUs    \\
		\midrule
		ABL1     & 123.5 (75.9)  & 59.2 (33.9) & 30.7 (17.2) & 21.2 (11.5) & 15.6 (8.5) & 14.3 (7.6) \\
		ACHE     & 93.6 (58.4)   & 53.7 (30.9) & 28.3 (14.5) & 17.4 (7.6)  & 14.1 (6.4) & 14.2 (6.5) \\
		ADAM17   & 101.9 (61.0)  & 43.8 (25.6) & 26.3 (17.0) & 14.8 (6.9)  & 11.8 (5.5) & 11.7 (5.4) \\
		ADORA2A  & 92.3 (58.2)   & 48.7 (31.3) & 26.1 (15.1) & 15.5 (8.1)  & 12.4 (6.7) & 12.3 (6.6) \\
		ADRB1    & 97.2 (62.2)   & 50.0 (31.8) & 30.6 (19.7) & 18.7 (10.3) & 14.7 (8.8) & 14.6 (8.7) \\
		ADRB2    & 104.1 (65.9)  & 53.7 (33.0) & 30.4 (18.3) & 19.3 (10.6) & 15.7 (9.2) & 15.7 (9.0) \\
		AKT1     & 74.8 (51.3)   & 57.1 (37.1) & 29.1 (18.4) & 18.1 (10.0) & 15.4 (8.9) & 15.1 (8.5) \\
		AKT2     & 95.9 (60.9)   & 48.5 (29.5) & 23.0 (12.4) & 16.7 (7.9)  & 13.0 (6.2) & 13.0 (6.2) \\
		AR       & 86.6 (55.8)   & 57.7 (36.4) & 31.1 (17.9) & 18.9 (9.8)  & 16.2 (9.0) & 15.7 (8.6) \\
		BACE1    & 88.4 (52.8)   & 48.5 (32.4) & 25.0 (15.2) & 13.7 (6.3)  & 11.7 (5.2) & 11.5 (5.1) \\
		CA2      & 100.2 (64.2)  & 56.7 (38.6) & 28.6 (16.9) & 17.0 (9.2)  & 14.5 (8.2) & 16.4 (9.2) \\
		CASP3    & 88.0 (54.6)   & 45.9 (28.4) & 27.5 (16.1) & 14.6 (7.4)  & 11.6 (5.8) & 11.7 (5.9) \\
		CDK2     & 68.9 (41.9)   & 44.6 (26.7) & 24.4 (13.7) & 14.2 (7.0)  & 11.6 (5.7) & 11.5 (5.6) \\
		CSF1R    & 91.2 (55.6)   & 44.2 (26.4) & 24.3 (13.5) & 14.8 (7.5)  & 11.7 (5.9) & 11.8 (5.8) \\
		CYP2C9   & 95.2 (59.4)   & 53.7 (31.6) & 29.7 (17.5) & 16.6 (8.2)  & 14.5 (7.6) & 15.7 (8.0) \\
		CYP3A4   & 85.9 (51.4)   & 48.6 (32.5) & 25.6 (15.6) & 15.3 (8.0)  & 12.8 (6.9) & 12.6 (6.9) \\
		DHFR     & 83.6 (50.5)   & 44.4 (26.8) & 23.6 (13.2) & 14.9 (7.7)  & 12.0 (6.2) & 12.0 (6.1) \\
		DPP4     & 103.3 (61.4)  & 60.0 (37.7) & 28.0 (14.6) & 16.6 (7.4)  & 13.6 (6.3) & 14.0 (6.4) \\
		DRD2     & 102.7 (68.4)  & 56.4 (39.2) & 29.8 (18.2) & 18.3 (10.1) & 16.1 (9.4) & 16.1 (9.2) \\
		DRD3     & 88.6 (59.8)   & 50.5 (36.3) & 26.4 (16.2) & 16.5 (8.8)  & 13.4 (7.2) & 13.5 (7.2) \\
		EGFR     & 83.8 (52.3)   & 50.0 (30.6) & 26.7 (14.8) & 16.6 (8.0)  & 13.4 (6.6) & 13.5 (6.6) \\
		ESR1     & 95.8 (58.0)   & 46.4 (27.4) & 25.6 (14.4) & 15.5 (7.7)  & 12.3 (6.3) & 12.3 (6.1) \\
		ESR2     & 89.7 (54.0)   & 47.1 (28.1) & 25.2 (13.4) & 15.3 (7.6)  & 12.2 (6.0) & 12.2 (6.0) \\
		F10      & 93.2 (56.8)   & 50.7 (37.8) & 25.2 (14.3) & 15.5 (7.8)  & 12.3 (6.0) & 12.2 (6.1) \\
		F2       & 89.5 (53.6)   & 46.4 (29.8) & 23.6 (13.3) & 13.2 (6.2)  & 11.3 (5.4) & 11.4 (5.3) \\
		FGFR1    & 285.8 (198.4) & 50.6 (33.8) & 27.7 (17.4) & 16.5 (9.1)  & 13.4 (7.7) & 13.4 (7.6) \\
		GBA      & 102.5 (65.9)  & 60.5 (43.4) & 30.7 (16.8) & 18.6 (9.1)  & 15.4 (8.1) & 15.4 (7.9) \\
		HMGCR    & 266.0 (165.0) & 47.8 (28.4) & 27.2 (14.7) & 16.0 (7.5)  & 12.7 (6.1) & 12.8 (6.1) \\
		HSD11B1  & 101.4 (64.8)  & 51.7 (37.0) & 27.7 (16.4) & 16.4 (8.6)  & 14.0 (7.4) & 13.9 (7.3) \\
		HSP90AA1 & 258.2 (159.9) & 47.0 (28.1) & 23.2 (13.2) & 16.4 (8.1)  & 13.1 (6.4) & 13.1 (6.4) \\
		IGF1R    & 91.5 (55.1)   & 45.8 (26.7) & 25.0 (13.6) & 15.7 (9.2)  & 12.2 (6.1) & 12.0 (6.0) \\
		\bottomrule
	\end{tabular}
\end{table}

\begin{table}
	\caption{
		Table~\ref{tab:docking-times} continued.
	}
	\label{tab:docking-times2}
	\begin{tabular}{lcccccc}
		\toprule
		Target   & 1 CPU         & 2 CPUs      & 4 CPUs      & 8 CPUs      & 16 CPUs    & 32 CPUs    \\
		\midrule
		JAK2     & 91.4 (56.7)   & 45.5 (27.5) & 24.9 (13.6) & 15.1 (7.4)  & 11.8 (5.9) & 11.8 (5.7) \\
		KDR      & 104.1 (64.5)  & 55.9 (37.3) & 26.9 (16.0) & 16.2 (8.4)  & 13.9 (7.3) & 14.0 (7.3) \\
		KIT      & 254.2 (157.7) & 46.0 (27.7) & 25.2 (14.2) & 15.4 (7.4)  & 12.4 (6.0) & 12.4 (5.9) \\
		LCK      & 85.2 (51.5)   & 42.4 (24.9) & 22.8 (12.4) & 14.0 (6.7)  & 11.1 (5.2) & 11.1 (5.1) \\
		MAOB     & 97.6 (59.1)   & 52.5 (30.8) & 36.1 (20.5) & 20.8 (10.5) & 17.2 (8.6) & 17.2 (8.5) \\
		MAP2K1   & 91.4 (58.8)   & 52.0 (31.8) & 28.3 (16.7) & 17.0 (8.8)  & 13.7 (7.2) & 13.5 (7.0) \\
		MAPK1    & 93.5 (58.3)   & 53.6 (38.3) & 25.2 (14.6) & 15.4 (8.1)  & 12.3 (6.2) & 12.5 (6.3) \\
		MAPK14   & 92.2 (58.8)   & 49.8 (39.4) & 25.7 (15.5) & 15.7 (8.7)  & 12.6 (7.1) & 12.6 (7.0) \\
		MAPKAPK2 & 88.3 (52.7)   & 53.7 (32.8) & 25.0 (14.0) & 15.2 (7.5)  & 11.9 (6.0) & 11.9 (5.8) \\
		MET      & 100.5 (62.3)  & 50.5 (30.0) & 27.5 (15.0) & 16.4 (7.7)  & 13.5 (6.5) & 13.3 (6.5) \\
		MMP13    & 86.2 (54.6)   & 49.2 (30.5) & 23.9 (13.8) & 14.2 (7.1)  & 11.3 (5.8) & 11.5 (5.8) \\
		NOS1     & 94.1 (62.0)   & 60.1 (39.7) & 28.9 (17.1) & 17.3 (8.5)  & 14.2 (7.1) & 14.4 (7.2) \\
		NR3C1    & 105.4 (65.2)  & 52.6 (32.5) & 30.5 (17.6) & 18.7 (9.5)  & 15.9 (8.8) & 15.7 (8.5) \\
		PARP1    & 82.3 (48.0)   & 48.4 (31.1) & 24.7 (13.3) & 14.7 (6.9)  & 11.7 (5.6) & 11.8 (5.5) \\
		PDE5A    & 85.2 (48.9)   & 51.0 (29.2) & 25.5 (13.6) & 15.2 (7.3)  & 12.1 (5.8) & 11.9 (5.7) \\
		PGR      & 89.0 (60.4)   & 61.7 (38.3) & 31.9 (19.5) & 19.1 (10.3) & 16.7 (9.5) & 16.1 (9.1) \\
		PLK1     & 93.7 (58.8)   & 49.7 (38.1) & 25.8 (14.7) & 15.5 (8.1)  & 12.4 (6.6) & 12.4 (6.4) \\
		PPARA    & 99.9 (63.1)   & 56.1 (34.7) & 27.4 (15.8) & 16.7 (8.6)  & 13.6 (7.4) & 13.7 (7.5) \\
		PPARD    & 89.0 (55.4)   & 52.9 (32.1) & 25.9 (15.1) & 16.1 (8.3)  & 13.7 (7.1) & 13.2 (6.8) \\
		PPARG    & 87.5 (54.1)   & 49.0 (29.8) & 23.8 (13.4) & 14.7 (7.0)  & 11.6 (5.6) & 11.6 (5.5) \\
		PTGS2    & 99.2 (61.1)   & 58.2 (35.5) & 29.7 (16.3) & 17.8 (8.4)  & 14.8 (7.4) & 14.6 (7.3) \\
		PTK2     & 256.9 (163.8) & 46.3 (29.0) & 25.5 (15.9) & 15.4 (8.3)  & 12.2 (6.6) & 12.3 (6.6) \\
		PTPN1    & 76.9 (46.9)   & 55.6 (36.1) & 25.6 (15.0) & 15.0 (7.0)  & 12.5 (6.1) & 12.1 (5.7) \\
		REN      & 85.9 (52.1)   & 43.8 (26.9) & 24.1 (12.8) & 14.6 (6.8)  & 11.7 (5.3) & 11.5 (5.3) \\
		ROCK1    & 69.3 (42.9)   & 52.8 (37.8) & 24.9 (14.2) & 14.6 (6.7)  & 16.3 (7.6) & 11.7 (5.4) \\
		SRC      & 90.8 (58.1)   & 47.3 (29.8) & 25.3 (15.2) & 14.5 (7.5)  & 12.5 (6.7) & 12.5 (6.5) \\
		THRB     & 101.5 (63.6)  & 55.2 (38.9) & 27.1 (15.6) & 16.0 (8.2)  & 14.4 (7.7) & 14.3 (7.6) \\
		\bottomrule
	\end{tabular}
\end{table}

%%%%%%%%%%%%%%%%%%%%%%%%%%%%%%%%%%%%%%%%%%%%%%%%%%%%%%%%%%%%
\subsection{Dataset}
\label{ssec:comp_details_dataset}
%%%%%%%%%%%%%%%%%%%%%%%%%%%%%%%%%%%%%%%%%%%%%%%%%%%%%%%%%%%%

Docking scores were computed in a cluster environment using the resources of the Cambridge Service for Data Driven Discovery (CSD3).
Each score was calculated using a single core of a Intel Xeon Skylake, 2.6GHz 16-core, on a node with 3.42MB of RAM.
In addition to the more than 15 million docking scores in the \dockstring{} dataset,
we also computed scores for assessing the quality of each target, and for determining the optimal search box sizes.
In total, the preparation and computation of the dataset required more than 500K CPU hours.

%%%%%%%%%%%%%%%%%%%%%%%%%%%%%%%%%%%%%%%%%%%%%%%%%%%%%%%%%%%%
\subsection{Baselines}
%%%%%%%%%%%%%%%%%%%%%%%%%%%%%%%%%%%%%%%%%%%%%%%%%%%%%%%%%%%%

The regression baselines were relatively inexpensive.
Each run of lasso, ridge regression, XGBoost, and \glspl{gp} took under 1h on a single machine with 6 CPUs.
MPNN and Attentive FP methods each took around 2h on a single machine with a NVIDIA 2080 Ti GPU.
Training of the virtual screening models was identical to the regression.
Prediction on ZINC took around 10 CPU hours for ridge regression and 1500 CPU hours for Attentive FP.
The molecular optimization tasks took between 24-72 hours to run for all methods (on a machine with 8 CPUs)
depending on the optimization trajectory and the number of calls to \dockstring{} required to evaluate each objective.
In total, we estimate that all benchmark tasks collectively required about 20k CPU hours.

%%%%%%%%%%%%%%%%%%%%%%%%%%%%%%%%%%%%%%%%%%%%%%%%%%%%%%%%%%%%
\section{Maintenance Plan}
\label{sec:maintenance}
%%%%%%%%%%%%%%%%%%%%%%%%%%%%%%%%%%%%%%%%%%%%%%%%%%%%%%%%%%%%

The \dockstring{} Python package will be hosted on GitHub and actively developed.
Since it is vital to ensure that the package is compatible with our dataset
(i.e.\@ that it can be used to generate the same numbers),
we will perform frequent testing to ensure that future changes do not change the numerical output of the package.
In particular, we will work to ensure that the package still functions even when new versions of the major dependencies
(i.e.\@ \texttt{rdkit}, \texttt{openbabel}) are released.
If breaking changes are required to implement new features, we will to split the project
into a different package (e.g.\@ \dockstring{}2) to preserve the original version.

The \dockstring{} dataset is fixed and will continue to be hosted on Figshare so that its standardized form can be accessed by researchers.
We are interested in expanding the dataset in the future to include more targets / ligands,
and plan to follow the model of ZINC \cite{irwin2012zinc,irwin2020zinc20}
by releasing updated versions of the dataset as separate re-numbered datasets (e.g.\@ \dockstring{}2022).

The code for \dockstring{} benchmarks will be hosted on GitHub upon publication.
We also plan to host a public leader board and list of publications that use \dockstring{}'s benchmarks.
The benchmarks presented in this work are only 3 of many possible benchmarks that \dockstring{} enables.
In future work, we plan to develop and promote other benchmarks, starting with tasks for transfer learning, meta-learning and few-shot learning.
These will likely be released with future publications and linked to on the \dockstring{} website.

\clearpage
\newpage

%%%%%%%%%%%%%%%%%%%%%%%%%%%%%%%%%%%%%%%%%%%%%%%%%%%%%%%%%%%%
% Graphical abstract
%%%%%%%%%%%%%%%%%%%%%%%%%%%%%%%%%%%%%%%%%%%%%%%%%%%%%%%%%%%%

% TOC figure: https://pubsapp.acs.org/paragonplus/submission/toc_abstract_graphics_guidelines.pdf

\begin{figure}[ht]
	\centering
	\includegraphics[width=3.25in]{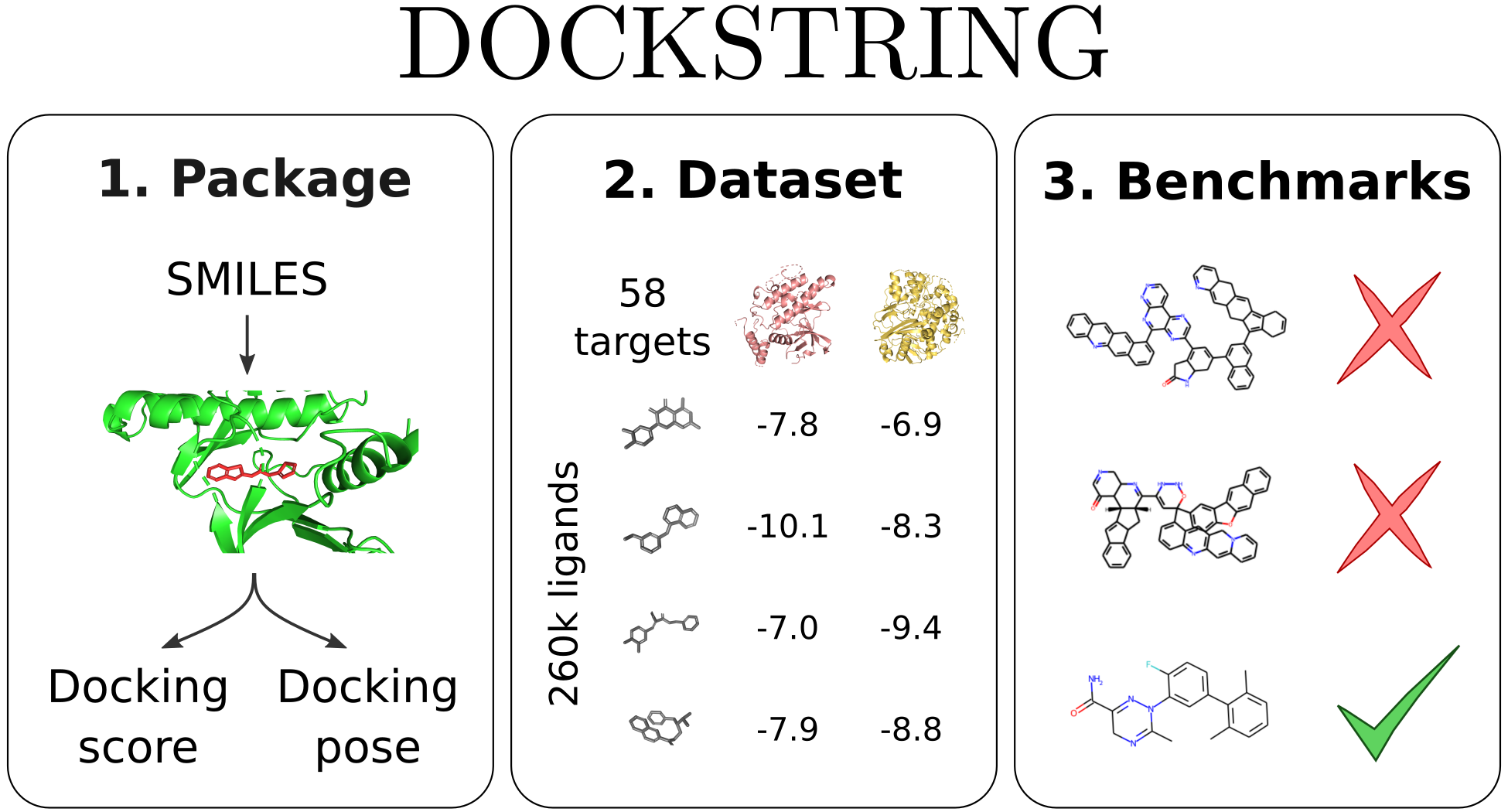}
	\caption{
	Table of Contents figure. 
	}
	\label{fig:toc}
\end{figure}

\end{document}